\documentclass[lettersize,journal]{IEEEtran}
\usepackage{amsmath,amsfonts}
\usepackage{algorithmic}
\usepackage{algorithm}
\usepackage{array}
\usepackage{booktabs}
\usepackage[caption=false,font=normalsize,labelfont=sf,textfont=sf]{subfig}
\usepackage{textcomp}
\usepackage{stfloats}
\usepackage{breqn}
\usepackage{graphicx}
\usepackage{mathtools}
\usepackage{url}
\usepackage{verbatim}
\usepackage{cite}
\usepackage{tkz-euclide}
\usepackage[percent]{overpic}
\usepackage{tikz}
\usepackage{pgfplots}

\newcommand\copyrighttext{%
  \footnotesize \textcopyright 2025 IEEE. Personal use of this material is permitted.
  Permission from IEEE must be obtained for all other uses, in any current or future
  media, including reprinting/republishing this material for advertising or promotional
  purposes, creating new collective works, for resale or redistribution to servers or
  lists, or reuse of any copyrighted component of this work in other works.}
\newcommand\copyrightnotice{%
\begin{tikzpicture}[remember picture,overlay]
\node[anchor=south,yshift=10pt] at (current page.south) 
  {\fbox{\parbox{\dimexpr\textwidth-\fboxsep-\fboxrule\relax}{\copyrighttext}}};
\end{tikzpicture}%
}
\usetikzlibrary{patterns}
\begin{document}

\title{LBONet: Supervised Spectral Descriptors for Shape Analysis}

\author{Oguzhan Yigit, Richard C. Wilson,~\IEEEmembership{Senior Member,~IEEE}
\thanks{Oguzhan Yigit and R. C. Wilson are with Department of Computer Science, University of York, UK.
E-mail: oguzhan.yigit@york.ac.uk, richard.wilson@york.ac.uk}
\thanks{Code available online under https://github.com/yioguz/LBONet}}


\maketitle
\copyrightnotice

\begin{abstract}
The Laplace-Beltrami operator has established itself in the field of non-rigid shape analysis due to its many useful properties such as being invariant under isometric transformation, having a countable eigensystem forming an orthornormal basis, and fully characterizing geodesic distances of the manifold. However, this invariancy only applies under isometric deformations, which leads to a performance breakdown in many real-world applications. In recent years emphasis has been placed upon extracting optimal features using deep learning methods, however spectral signatures play a crucial role and still add value. In this paper we take a step back, revisiting the LBO and proposing a supervised way to learn several operators on a manifold. Depending on the task, by applying these functions, we can train the LBO eigenbasis to be more task-specific. The optimization of the LBO leads to enormous improvements to established descriptors such as the heat kernel signature in various tasks such as retrieval, classification, segmentation, and correspondence, proving the adaptation of the LBO eigenbasis to both global and highly local learning settings.
\end{abstract}


\section{Introduction}
\IEEEPARstart{S}{hape} analysis has been greatly affected by the recent introduction of deep learning techniques, resulting in numerous papers on the incorporation of machine learning methods in shape analysis tasks. When looking at these papers chronologically, we can see that the state-of-the-art is a result of accumulative work. Most of the classical supervised and unsupervised learning methods are based on Euclidean data. However, graphs and Riemannian manifolds which represent shape are non-Euclidean, we cannot directly apply these methods to them. Hence, the primary focus of research in shape analysis in recent years was to find analogies of these methods in the non-Euclidean domain, such as the \textit{patch} operator \cite{kokkinos2012intrinsic}, supervised bag of visual words \cite{litman2014supervised}, and non-Euclidean convolution \cite{masci2015shapenet}. These efforts led to the successful application of convolutional neural networks and deep learning techniques on manifolds, both in the spatial and spectral domain. Both domains have their unique features, but a significant drawback of spectral CNNs has always been the lack of generalization across manifolds, even though techniques, such as synchronized Spectral CNN \cite{yi2017syncspeccnn} were introduced to overcome this. 

\begin{figure}[!th]
\centering
\begin{tikzpicture}[scale=0.8]


\coordinate (A) at (0.4,0.2); 
\coordinate (B) at (3,0);
\coordinate (C) at (1.5,2);
\coordinate (D) at (3.6,1.8); 

\draw[pattern=north west lines, pattern color=gray!50] (A) -- (B) -- (C) -- cycle;
\draw[pattern=north west lines, pattern color=gray!50] (C) -- (D) -- (B);

\draw (A) -- (B) -- (C) -- cycle;
\draw (B) -- (D) -- (C);
\draw [line width=0.5mm, red] (C) -- (B);

\node[text width=4cm, align=center] at (2,-1) {Riemannian metric weighting (RiemannNet)};
\end{tikzpicture}
\hspace{-1cm}
\begin{tikzpicture}[scale=0.8]
\begin{scope}[shift = {(0.3, -0.2)}]

\coordinate (A) at (0.6,1.1);
\coordinate (B) at (1.35,0.7);
\coordinate (C) at (2.3,1.2);
\coordinate (D) at (1.2,2.2);
\coordinate (E) at (0.6,1.8);
\coordinate (F) at (1.8,2.3);
\coordinate (G) at (2.4,1.9);
\draw[red, line width=0.5mm, pattern=north west lines, pattern color=gray!50] (E) --(A) -- (B) -- (C) -- (G) -- (F) -- (D) -- cycle; 

\coordinate (A) at (0.5,0.2); 
\coordinate (B) at (2.3,0.2);
\coordinate (C) at (1.4,1.7);
\coordinate (D) at (3.3,1.5); 
\coordinate (E) at (2.8,2.5);
\coordinate (F) at (1.5,2.6);
\coordinate (G) at (0.5,2.3);
\coordinate (H) at (0,1.3);
\draw (A) -- (B) -- (C) -- cycle;
\draw (C) -- (D) -- (B) -- cycle;
\draw (C) -- (D) -- (E) -- cycle;
\draw (C) -- (F) -- (E) -- cycle;
\draw (C) -- (F) -- (G) -- cycle;
\draw (C) -- (H) -- (G) -- cycle;
\draw (C) -- (H) -- (A) -- cycle;

\draw (A) -- (B) -- (C) -- cycle;
\draw (B) -- (D) -- (C);


\end{scope}
\node[text width=4cm, align=center] at (2,-1) {Voronoi cell weighting (VoronoiNet)};
\end{tikzpicture}
\hspace{-1cm}
\begin{tikzpicture}[scale=0.8]

\coordinate (A) at (0.4,0.2); 
\coordinate (B) at (3,0);
\coordinate (i) at (0,1);
\coordinate (j) at (1.9,2.2);
\coordinate (k) at (1.65,0.74);
\coordinate (C) at (1.5,2);
\coordinate (D) at (3.6,1.8); 

\draw[pattern=north west lines, pattern color=gray!50] (A) -- (B) -- (C) -- cycle;
\draw[pattern=north west lines, pattern color=gray!50] (C) -- (D) -- (B);


\draw (A) -- (B) -- (C) -- cycle;
\draw (B) -- (D) -- (C);

\draw[dashed,-{Stealth[black!60]}, black!60] (barycentric cs:A=1,B=1,C=1) -- (0,1);
\draw[dashed,-{Stealth[black!60]}, black!60] (barycentric cs:A=1,B=1,C=1) -- (1.9,2.2);

\coordinate (i) at (0.9,2);
\coordinate (j) at (2.9,1.5);
\coordinate (k) at (1.65,0.74);

\draw[-{Stealth[red!70]}, red!70,line width=0.5mm] (barycentric cs:A=1,B=1,C=1) -- (1.1,1.7);
\draw[-{Stealth[red!70]}, red!70,line width=0.5mm] (barycentric cs:A=1,B=1,C=1) -- (3.2,1.7);

\coordinate (i) at (0,1);
\coordinate (j) at (0.9,2);
\coordinate (k) at (1.65,0.74);

\pic [draw,line width=0.5mm,red, angle radius=0.4cm]{angle = j--k--i};

\coordinate (i) at (1.9,2.2);
\coordinate (j) at (2.9,1.5);
\coordinate (k) at (1.65,0.74);

\pic [draw,line width=0.5mm,red, angle radius=0.4cm]{angle = j--k--i};

\node[text width=4cm, align=center] at (2,-1) {Anisotropy direction and factor (ALBONet)};

\end{tikzpicture}
\caption{LBONet comprises multiple modules which operate directly on the input mesh. RiemannNet is able to weight Riemannian metric on a manifold, ALBONet introduces anisotropy by rotating and factoring the diffusion speed, and lastly VoronoiNet weights the Voronoi cells. These affect the Laplace-Beltrami operator in several different ways, allowing the operator to adapt to the given data and alleviate the performance breakdowns incurred by non-isometric deformations.}
\label{fig:human}
\end{figure}
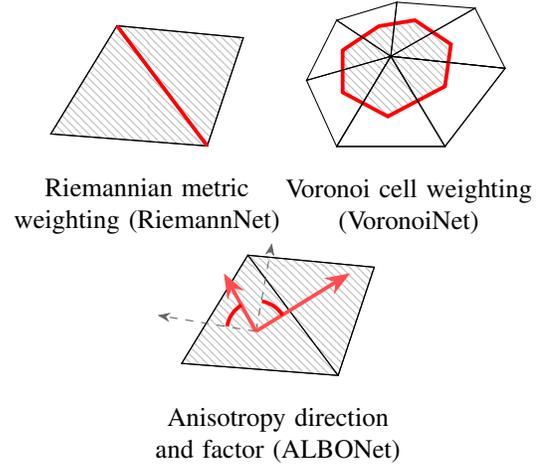
Almost forgotten by the focus on global encoding techniques, recent frameworks PointNet++ \cite{qi2017pointnet++}, second-order spectral transform \cite{yu2020second}, DiffusionNet \cite{sharp2022diffusionnet} demonstrate the continuing importance of spectral descriptors for non-rigid shape analysis, where a significant increase of accuracy is achieved on challenging benchmarks by using spectral descriptors as opposed to using only geometric properties such as point coordinates. 

Although the performance gain using spectral descriptors is becoming less as recent works have shown. The presented work can be plugged in any existing pipeline making use of spectral descriptors and promises to achieve further gains, which could make all spectral-based methods more attractive in the wider literature.
This not only proves the capability of feature retrieval networks to make use of spectral descriptors, but also their descriptive power. Deep learning methods slowly superseded dictionary-based retrieval methods such as bag of words \cite{litman2014supervised} or Fisher and Super Vector \cite{limberger2015feature}, due to their ease of use and end-to-end learning capability, however it is still not clear how they compare to those methods since \cite{giachetti2012radial}\cite{ye2013fast}\cite{limberger2018curvature} have shown good performance in common benchmarks. Many papers have been published on modifying the LBO eigenbasis \cite{limberger2018curvature}\cite{choukroun2018hamiltonian}\cite{andreux2015anisotropic}, but very few papers dedicate themselves learning through the LBO eigenbasis. Some attempts have been made recently such as HodgeNet\cite{smirnov2021hodgenet}, but competing methods have overcome them quickly. However backpropagating through the LBO eigendecomposition remains challenging and despite dealing with sparse matrices the backpropagation involves dense matrices, which in turn introduce problems with scalability. This work introduces custom PyTorch routines to deal with the given inherent sparsity of the data, to speed up the backpropagation, allowing to process high resolution data in a reasonable timeframe.

This paper takes an unconventional way and dedicates itself on how learning can be enabled through the LBO eigenbasis, in order to improve spectral descriptors and to make them more task-specific. 

As demonstrated by this work spectral methods can beat state-of-the-art methods after weighting the LBO accordingly. The learned weights roughly reflect the task at hand, e.g. in a segmentation task, boundary regions express increased activity, whereas in a correspondence tasks, near-isometric deformations on the manifold express increased activity, making it easier for descriptors or MLPs to pickup the right details from the spectral descriptor. The intriguing question this paper tackles is how much learning potential is hidden in the LBO eigenbasis, how much of it can be exploited, and what are the limiting factors to achieve further performance gains.

These performance gains are achieved by introducing weighting layers, which we call modules in the following, which weight the manifold in different ways and modify the nature of the LBO. In prior work it was not possible to learn them with backpropagation and methods have either used heuristics or precalculation of many combinations of parameters. Learning these modules offers more advantages over the traditional methods as the learning can learn complex non-linear relationships in the data and is not restricted by predefined values. LBONet enables the intrinsic learning of features, which is not possible when relying solely on extrinsic information such as raw coordinates, an approach that has received significant emphasis in recent literature.
The ablation study shows that LBONet is able to outperform some methods by just weighting the LBO and producing a spectral descriptor. It produces competitive results, when paired with a backend, which utilizes the intrinsic gains of LBONet.

\textbf{Contributions}
We present an end-to-end learning framework for learning operators on triangle meshes to improve the performance of spectral descriptors. With the introduction of our framework we make the following contributions,

\begin{itemize}
  \item a learnable block which learns a Riemannian metric weighting layer based on intrinsic shape features
 \item a learnable block which learns Voronoi cell weighting  layer based on intrinsic shape features;
      \item a learnable block which learns an anisotropic LBO (ALBO) layer based on intrinsic shape features and optionally into two directions (ALBO+)
  \item a modern backpropagation implementation for differentiating eigenderivatives in PyTorch, which exploits the the sparsity of the data and addresses only as few eigenfrequencies as set;
  \item an end-to-end approach for learning, starting from features on the mesh up to the descriptor;
  
\end{itemize}

\section{Background}
\label{chap:back}
\begin{figure}[H]
\centering
\begin{tikzpicture}

\coordinate (A) at (0.6,1.1);
\coordinate (B) at (1.35,0.7);
\coordinate (C) at (2.3,1.2);
\coordinate (D) at (1.2,2.2);
\coordinate (E) at (0.6,1.8);
\coordinate (F) at (1.8,2.3);
\coordinate (G) at (2.4,1.9);
\draw [red, pattern=north west lines, pattern color=blue!30] (E) --(A) -- (B) -- (C) -- (G) -- (F) -- (D) -- cycle; 
\foreach \point/\position in {A/, B/, C/, D/, E/, F/, G/} {
    \fill (\point) circle (0.5pt);
}

\coordinate (A) at (0.5,0.2); 
\coordinate (B) at (2.3,0.2);
\coordinate (C) at (1.4,1.7);
\coordinate (D) at (3.3,1.5); 
\coordinate (E) at (2.8,2.5);
\coordinate (F) at (1.5,2.6);
\coordinate (G) at (0.5,2.3);
\coordinate (H) at (0,1.3);
\draw (A) -- (B) -- (C) -- cycle;
\draw (C) -- (D) -- (B) -- cycle;
\draw (C) -- (D) -- (E) -- cycle;
\draw (C) -- (F) -- (E) -- cycle;
\draw (C) -- (F) -- (G) -- cycle;
\draw (C) -- (H) -- (G) -- cycle;
\draw (C) -- (H) -- (A) -- cycle;

\draw (A) -- (B) -- (C) -- cycle;
\draw (B) -- (D) -- (C);

\foreach \point/\label in { D/$k_2$}
    \fill (\point) circle (2pt) node[right] {\label};
\foreach \point/\label in {C/$i$}
    \fill (\point) circle (2pt) node[xshift=-0.1cm, above] {\label};
\foreach \point/\label in {B/$j$}
    \fill (\point) circle (2pt) node[below] {\label};
\foreach \point/\label in {A/$k_1$}
    \fill (\point) circle (2pt) node[left] {\label};

\foreach \point/\label in { E/, F/, G/, H/}
    \fill (\point) circle (1pt);

\pic["$\alpha_2$", draw=black, angle radius=0.7cm] {angle = C--D--B};
\pic["$\alpha_1$", draw=black, angle radius=0.7cm] {angle = B--A--C};

\foreach \point/\position in {A/, B/, C/, D/, E/, F/, G/} {
    \fill (\point) circle (0.5pt);
}
\end{tikzpicture}
\caption{Voronoi cell on vertex $i$, which is outlined in red. The Voronoi cell can be scaled and has only impact on the local Voronoi cell.}
\label{fig:voronoi}
\end{figure}
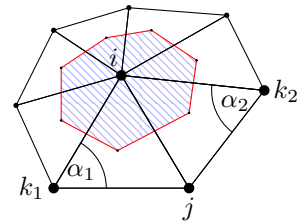

\paragraph{Laplace-Beltrami Operator}
The Laplace-Beltrami operator (LBO) is a generalization of the classical Laplace operator to Riemannian manifolds. Given a twice-differentiable real-valued function $f \in C^2$ defined on a manifold $M$, the LBO is defined as:

\begin{equation}
\Delta f = \operatorname{div} (\nabla f),
\label{eq:LBOFIRST}
\end{equation}
where $\nabla f$ denotes gradient of $f$ and $\operatorname{div}$ the divergence on the manifold.
The LBO was first introduced to shape analysis through ShapeDNA \cite{reuter2005laplace}, which uses the eigenvalues of the operator on a manifold to construct an isometry-invariant spectral embedding of the shape.
\\
\textbf{Spectral Theorem}
The LBO is a self-adjoint, semi-positive definite operator and therefore admits an orthonormal eigensystem, which can be retrieved by solving the eigenvalue problem:
\begin{equation} \label{eq:eig}
\Delta \phi_k = \lambda_k \phi_k,
\end{equation}
where $\lambda_k$ and $\phi_k$ is the corresponding pair of eigenvalue and eigenvector, respectively.
As we deal with discrete approximations of manifolds such as meshes, point clouds, and voxels, we need to discretize the LBO. Several discretization methods have been proposed for triangle meshes, but the cotangent weight method  \cite{meyer2003discrete} has established itself, and is defined as:
\begin{equation}
\Delta f = Lf = A^{-1}Wf,
\end{equation}
where $L$ denotes the discrete Laplacian, and $W$ is referred to as \textit{stiffness matrix} and its elements are defined by:
\begin{equation}
W(i,j) = \begin{cases}
    \frac{\cot \alpha_{1}+\cot \alpha_{2}}{2}, & \text{if }(i,j) \in E\\
    - \sum_{k \neq i} W(i,k), & \text{if }i=j\\
0, & \text{otherwise},
\end{cases}  
\label{eq:stiffness}
\end{equation}
$\alpha_1$ and $\alpha_2$ are the opposing angles in an edge flap, which is depicted in Figure \ref{fig:edgeflap}.
 $A$ is referred to as  \textit{mass matrix} and is a diagonal matrix and is composed of the area of the \textit{Voronoi} cell,
 \begin{equation}
A(i,j) = \begin{cases}
    \frac{1}{8} \sum_{j \in N_1(i)}(\cot \alpha_{1}+\cot \alpha_{2}) || d_{ij} ||^2, & \text{if }i=j\\
0, & \text{otherwise},
\end{cases}  
\end{equation}

where $d_{ij}$ denotes the distance between vertex $i$ and $j$, and $N_1(i)$ is the one-ring neighbourhood of vertex $i$ (see Figure \ref{fig:voronoi}).
Another popular choice for the mass matrix is using barycentric cells \cite{sharp2022diffusionnet} \cite{gahm2018riemannian}.
Finally, we derive the generalized eigenvalue problem:
\begin{equation}
W\phi = \lambda A \phi.
\label{eq:generalized}
\end{equation}
Following ShapeDNA's \cite{reuter2005laplace} example, Rustamov \cite{rustamov2007laplace} introduced the global point signature (GPS), which is defined at vertex $i$ as,
\begin{equation}
GPS(i) = \left(\frac{1}{\sqrt{\lambda_{1}}}\phi_{1}(i), \frac{1}{\sqrt{\lambda_{2}}}\phi_{2}(i), \dots, \frac{1}{\sqrt{\lambda_{n}}}\phi_{n}(i)\right),
\end{equation}
where $n$ defines the upper boundary for the eigensystem.
Despite being the first signature making use of the full spectrum of the LBO, it suffered from several drawbacks such as the flipping of eigenvectors due to said eigenvectors being defined only up to a sign and swapping of eigenvectors. Motivated by aforementioned drawbacks, Sun \textit{et al}. \cite{sun2009concise} introduced a signature based on the heat diffusion process, which is still relevant in most shape analysis methods. The heat diffusion process is given by
\begin{equation}
\Delta f(i,t) = - \frac{\partial f (i,t)}{\partial t},
\end{equation}
where $f(x, t)$ is the temperature at point $i$ at time $t$. By the \textit{informative theorem} \cite{sun2009concise}, the heat kernel can be expressed in the spectral domain by restricting it to the diagonal elements:
\begin{equation}
\label{eq:hks}
h_{t} (i, i) = \sum_{k>0} e^{-t \lambda_{k}} \phi_{k}(i)^{2}.
\end{equation}
As the heat kernel varies with the scale of a manifold, which can cause some performance breakdowns, a scale-invariant variant version was proposed by Bronstein \textit{et al.} \cite{bronstein2010scale}.
\section{Related Work}\label{chap:rel}
As the term \textit{spectral methods} is used interchangeably for methods based on spectral descriptors and methods which aim to improve the underlying LBO by changing its characteristics, it is crucial to draw a contrast between them since they tackle the same problem in completely different ways. In the following, we therefore refer to those methods based on spectral descriptors as \textit{spectral methods} and to methods which manipulate the behaviour of the underlying LBO as \textit{LBO methods}.
\subsection{Spectral Methods}
The introduction of the heat and wave kernel signature has spawned a plethora of methods in an attempt to overcome the downsides of spectral descriptors.
Although these drawbacks are intrinsic to the LBO itself, learning from spectral descriptors can help mitigate some of them. There are heuristic methods which optimize the hyperparameters of these signatures such as scaling the eigenvalues in improved wave kernel signature \cite{limberger2015feature} or optimizing the variance parameter in the wave kernel signature \cite{naffouti2018heuristic}. Litman \textit{et al}. parameterize spectral descriptors \cite{litman2013learning} and make it possible to systematically learn an optimal signature by metric learning.
Other methods on the other hand rely completely on the descriptive power of spectral descriptors and learn with an elaborate feature retrieval network. PointNet++ \cite{qi2017pointnet++} introduced a novel method to consume point clouds and applied it to the non-rigid domain by using spectral descriptors and geometric features as input. Second-order spectral transform \cite{yu2020second} employed second-order pooling on spectral descriptors and based their learning on top of the second-order pooled descriptor and  learn on the resulting SPDM-manifold. They achieve moderate results on SHREC'15 but achieve unparalleled performance on the challenging SHREC'14 human benchmarks, which to the best of our knowledge, is outperformed only by our method. Dynamic graph convolutional neural network \cite{wang2019dynamic} constructs a local neighborhood graph and learns a neighbourhood function, allowing the sharing of information across k-nearest neighbours, thereby adding an additional layer for learning. DiffusionNet \cite{sharp2022diffusionnet} takes optionally spectral descriptors as input features and applies spatially shared MLPs on a feature construct, but does not learn in the spectral domain itself.

\subsection{LBO Methods}
As the downsides of spectral signatures are inherent to the LBO, the question arises: to what degree can these downsides be compensated for by modifying the nature of the LBO. The kinetic Laplace-Beltrami operator \cite{limberger2018curvature} achieves significant boost of performance by weighting the mass matrix with a kinetic energy term derived from the curvature of the manifold. Similarly, the Hamiltonian \cite{choukroun2018hamiltonian} operator applies an optimized potential $V$ to the spectrum and thus changes the behaviour of the LBO by modifying the diagonal entries of the stiffness matrix using perturbation theory.

Previously, backpropagating through the LBO eigenbasis was non-trivial, methods either relied on heuristics or precalculation with a set of different parameters such as anisotropic diffusion descriptors \cite{boscaini2015learning} or anisotropic convolutional neural network \cite{boscaini2016anisotropic}. Another intriguing extension of the ALBO is the FLBO \cite{weber2024finsler}, which removes the quadratic assumption of the Riemmanian metric, allowing asymmetries on the manifold. Synchronized Spectral CNN \cite{yi2017syncspeccnn} applies a spectral transform network on the LBO eigenbasis, successfully generalizing the learned coefficients—a significant advancement in extending the learning capabilities of the LBO. We compare our results on ShapeNet against this method, as both operate at the LBO level. Other notable methods include the Learned Binary Spectral Shape Descriptor \cite{xie2016learned}, which leverages LBO eigenvectors within a metric network to achieve superior performance in correspondence tasks, and DeepShape \cite{xie2015deepshape}, which trains an autoencoder on extracted HKS features from manifolds to create a highly effective shape descriptor for downstream applications.

Some methods however attempted successfully to backpropagate through the LBO eigenbasis, one of which is \cite{cosmo2019isospectralization}, where a shape is iteratively optimized wrt. the eigenvalues of a target shape and then is used in follow up tasks such as finding correspondences between these "isospectralized" shapes. Isospectralization is a concept, which aims to to reduce the spectral distance of  shapes as much as possible eventually resulting in a similar spectrum, hence the coined term Isospectralization. A similar work has been presented in \cite{rampini2019correspondence}, where this concept has been incorporated to Hamiltonian spectrum and partial shapes. Although isospectralization is an effective tool, recent works achieve high accuracies in correspondence tasks with higher degrees of non-isometry, proving that correspondences can be learnt even if the eigenbasis is not aligned.
The presented work can also be seen as a part of Isospectralization, as LBONet aims to a task-specific LBO eigenbasis, which in a correspondence setting leads to a spectrum, which is as close as possible to similar manifolds. As the experiments show, Isospectralization is only one of many things LBONet is able to achieve. In a segmentation task it is able to spread the segmented parts in the spectrum. Furthermore, the presented work operates on both levels eigenvalues and eigenvectors.

The methods introduced by conformal metric optimization on surfaces \cite{shi2011conformal} and Riemannian metric optimization on surfaces \cite{gahm2018riemannian}, despite their limited application to intrinsic brain mapping, have been groundbreaking as they form the first rudimentary end-to-end optimization of spectral descriptors. In the former, the authors optimize the mass matrix, whereas in the latter, the authors optimize the Riemannian metric, which translates to optimizing both, stiffness and mass matrix. More recently, HodgeNet \cite{smirnov2021hodgenet} follows the same goal of learning end-to-end by backpropagating through the eigenbasis of the \textit{Laplace-Beltrami} operator by approximating the gradients wrt.\ Hodge star operators.

The presented paper instead calculates the exact gradients, which do not require the calculation of additional eigenvectors such as in \cite{smirnov2021hodgenet}. Furthermore, it introduces several learning operators, which were previously only based on heuristics or precalculation, and allows to learn optimal weights for a given dataset.

\section{Approach}
\label{chap:approach}
\subsection{Building blocks}
\begin{figure}[h!]
    \includegraphics[scale=0.2, trim=8cm 2cm 1cm 2cm]{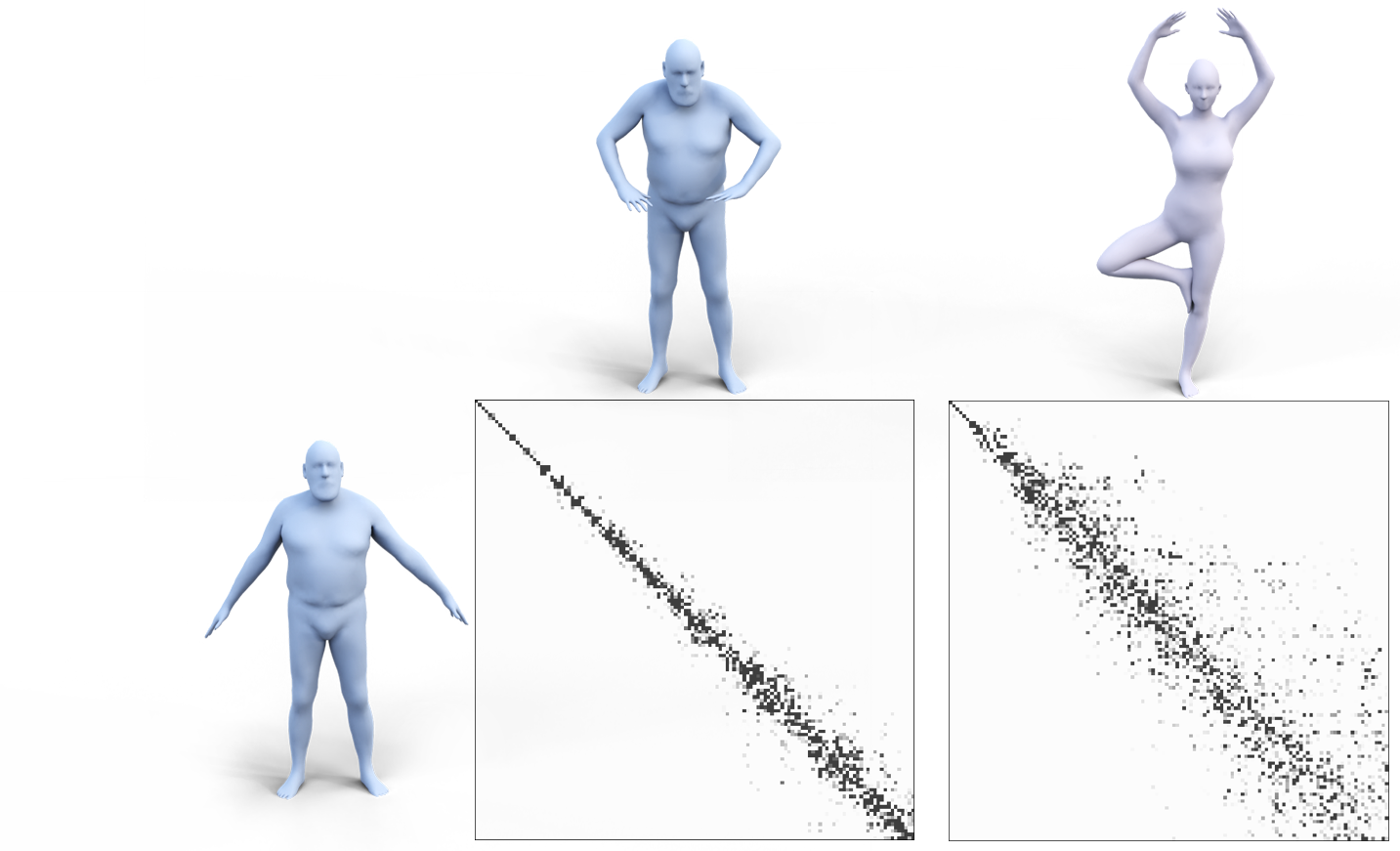}
    \caption{Functional matrix between an intra-class manifold pair (left blue and top purple) and a inter-class manifold pair (left blue and top purple). One can observe two things, first, the functional matrix deviates from the diagonal even with mild non-isometries and second, that with greater deformation, the matrix deviates even further. } 
    \label{fig:eigbasiscomparison}
\end{figure}

The following approach is based on the fact that any non-isometric deformation distorts the LBO eigenbasis drastically. This can be illustrated by using functional maps \cite{ovsjanikov2012functional}, where the C matrix, which maps the eigenbases on one manifold to another, deviates from the diagonal matrix with even mild near-isometries. The greater the non-isometric deformation, the further the map deviates from the diagonal, making the map more complex and less sparse and due to this fact the LBO eigenbasis become less ideal as a basis function. This effect is illustrated in Figure \ref{fig:eigbasiscomparison}, with an intra-class manifold for near-isometry and an inter-class manifold for non-isometry.
This sensitivity and instability of the LBO eigenbasis introduces two drastic downsides, firstly, it makes it difficult to transfer features learnt in the spectral space and, secondly, the overlying descriptor needs to span the spectral range a manifold can be embedded in, which could lead to a less precise descriptor.

To alleviate this, learning can be done both implicitly and explicitly. Implicit learning would learn an optimal descriptor based on the LBO eigenbasis and learning thereby only implicitly an ideal LBO eigenbasis, e.g. for a segmentation task, the learning can be done on the heat kernel directly, which then propagates the loss back to the eigenbasis. Whereas explicit learning would learn an LBO eigenbasis directly from the loss function and the loss would be given directly on the LBO eigenbasis, e.g. in a correspondence task, we might learn the eigenbasis directly, rather than going via an overlying descriptor.

The presented approach and experiments are all done by learning the LBO eigenbasis implictly. Learning the LBO eigenbasis explicitly is beyond the scope of the presented paper and is subject to further research.

\begin{figure*}[h]
\centering
\includegraphics[scale=0.92,trim=4.2cm 0.4cm 0cm 0.7cm]{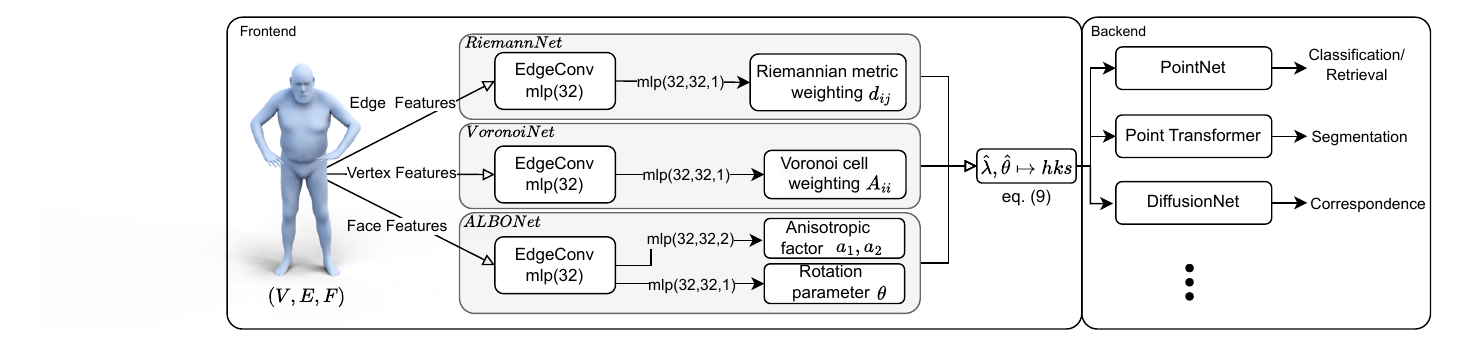}
\caption{\textbf{LBONet Architecture.} LBONet takes features on mesh edges, vertices, and faces as input of its three building blocks, RiemannNet, ALBONet and VoronoiNet. Each block consists of one EdgeConv which is creating a local neighbourhood graph as in \cite{wang2019dynamic} and learns an neighbourhood function. Then the features are passed through multiple shared MLPs to predict either the Riemannian metric on an edge, the anisotropy on a face or the Voronoi cell weighting at a vertex. These predictions are applied directly on the discretized LBO, modifying the stiffness and mass matrix, which we detail out from subsection D onwards. Its eigenbasis is then used to form a spectral descriptor, which then can be passed into a "backend" for further processing.}
\label{fig:arch}
\end{figure*}

We give the gradients in the following in the context of learning through the heat kernel signature, but they can be analogously derived for any other spectral signature such as the wave kernel signature \cite{aubry2011wave} or average mixing kernel signature\cite{cosmo20223d}. With the given loss LBONet takes the aim to weight the LBO s.t. the produced eigenbasis and the descriptor is better at the given task. Weighting the LBO is a common practice and as described in the related work sections used quite effectively in various tasks. However, as the backpropagation through the LBO eigenbasis is a challenging task, hyperparameters could not be learned easily and were therefore based on heuristics. We show the weighting based on three different elements, which are depicted in gray in Figure \ref{fig:arch}. Firstly, by using features on mesh edges to weight the Riemannian metric (RiemannNet), which by default is defined as the edge length of the mesh. Secondly, on faces, by adjusting the anisotropy (ALBONet), and lastly using features on vertices to weight the Voronoi cell (VoronoiNet).

We equip each block with EdgeConv which creates a local neighbourhood graph as in \cite{wang2019dynamic} and learns a neighbourhood function. This allows the sharing of information in the local neighborhood and to learn more effectively generalizing features. This proved far superior to learning directly from the features only sourcing their information from the local context. As a distance function, Euclidean distance showed good performance, although in some cases constructs neighbourhood erroneously, e.g. when a disjoint body part such as a hand is close to the hip, the hip might exchange information with the hand. We observed some downside effect in the experiments, however, the drawback was negligible, hence we omitted the construction of a geodesic neighborhood.

These features are then passed through multiple shared MLPs to predict the Riemannian metric weighting, which can shorten or expand distances on specific regions by scaling the Riemannian metric, steer the anisotropy individually on each face by rotating the orthogonal prinicpal curvature basis and scaling the anisotropic factors, and finally weight the Voronoi cell differently across the manifold. These predictions are applied directly on the discretized LBO, modifying the stiffness $L$ and mass matrix $A$ in various ways, which we detail out from subsection D onwards. This modified LBO eigenbasis  $(\hat{\lambda}, \hat{\phi})$ is then used to evaluate a spectral descriptor, which then can be either passed into a "backend" (see Figure \ref{fig:arch} for further processing or used directly for various tasks.
\subsection{Feature Learning}
Works on learning from unordered point clouds such as in PointNet \cite{qi2017pointnet}, PointNet++ \cite{qi2017pointnet++}, and DGCNN \cite{wang2019dynamic} or spectral methods such as Spectral Transform Block \cite{yu2020second} proved useful for shape analysis tasks. More recently point (cloud) transformers \cite{guo2021pct} \cite{wu2022point} and the heat kernel based transformer \cite{wong2023heat} have further pushed the boundaries in extracting powerful features.

In particular, their performance in non-rigid shape analysis using spectral descriptors rather than point coordinates proves that they are able to utilize the information within spectral descriptors, and furthermore, that there is still unexploited potential in them, as the standard LBO is not invariant under non-isometric deformation. The presented work explores to what extent spectral descriptors can contribute to the descriptiveness of features when using state-of-the-art feature extracting networks, by putting as much learning as possible into the spectral descriptor and the LBO eigenbasis. We highlight the tradeoffs of this level of adaption of the LBO to the given task in the conclusion.

\subsubsection{Edge-, Face-, Vertexwise Perceptrons}
With the propagated loss through the LBO eigenbasis, we can learn features from mesh edges by RiemannNet, from mesh faces by ALBONet and from mesh vertices by VoronoiNet. For this we use the concept of shared MLPs, which has been used widely in the literature and been referred to as *-wise perceptrons. Furthermore, experiments have shown that EdgeConv introduced by \cite{wang2019dynamic} yielded an additional performance gain. For edges and faces, we take the mean of their coordinates to build a coordinate system to find the nearest neighbours and use the asymmetric edge function given by \cite{wang2019dynamic},
\begin{equation}
    h_\Theta(x_i, x_j) = h_\Theta(x_i, x_j-x_i).
\end{equation}
followed by several shared MLP layers.
\subsubsection{Input Features}
As input to our pipeline we use intrinsic shape features derived from the second fundamental form such as mean curvature \cite{rusinkiewicz2004estimating}, gaussian, shape index, curvature index. For faces and edges we interpolate the features from their vertices. The intention is to limit the input features to intrinsically invariant features to avoid any dependency on the shape embedding and also have robust features, which are mostly independent from the meshing structure and resolution.

Usage of point coordinates as input features could yield an additional benefit, however at the cost of extensive training. Furthermore, intrinsic shape features prove to be significant for non-rigid shape analysis as shown in \cite{tang2012evaluation} and are widely used as enhancing input features \cite{qi2017pointnet++} \cite{wang2019dynamic} \cite{sharp2022diffusionnet} for non-rigid shape analysis.

The predictions of the the modules (RiemannNet, ALBONet, and VoronoiNet) modify the stiffness and mass matrix in different ways, which then can be used to solve the generalized eigenvalue problem,
\begin{equation}
    W \hat{\phi} = \hat{\lambda} A \hat{\phi}
\end{equation}
resulting in a modified task-specific eigenbasis $(\hat{\lambda}, \hat{\phi})$.

\subsection{Sensitivity Analysis of Spectral Descriptors}

To learn an optimal LBO eigenbasis, one needs to differentiate the generalized eigenvalue problem (\ref{eq:generalized}) wrt.\ to a given learnable parameter $p$ of the manifold, where $p$ could, for example, represent the edge length of a triangle:
\small
\begin{equation}
    \frac{\partial W}{\partial p_{ij}} \phi + W \frac{\partial \phi}{\partial p_{ij}} = \frac{\partial \lambda}{\partial p_{ij}}A\phi + \lambda \frac{\partial A}{\partial p_{ij}}\phi + \lambda A \frac{\partial\phi}{\partial p_{ij}}.
    \label{eq:generalizedNelson}
\end{equation}

\normalsize

Differentiating the stiffness $W$ and mass matrix $A$ is trivial, whereas the eigenbasis involves multiple steps. Nelson's method \cite{nelson1976simplified} has established itself for differentiating eigenvalues and eigenvectors and has been used widely throughout the literature. For the mass matrix, rather than using barycentric cells as in \cite{shi2011conformal}\cite{gahm2018riemannian}, we opt for using Voronoi cells introduced by \cite{meyer2003discrete}, as they prove to be more robust to meshing structure and is recommended by \cite{meyer2003discrete}, due to having a provably tight error bound.
Depending on whether the task at hand impacts the stiffness and/or mass matrix, Nelson's method needs to be adjusted accordingly. In the following we demonstrate the default method, but make use of all variations of Nelson's method, stiffness \& mass matrix, stiffness only, and mass only in the following sections.

From equation \ref{eq:generalizedNelson} we obtain for eigenvalue $\lambda_k$,
\small
\begin{equation}
    \frac{\partial \lambda_{k}}{\partial p_{ij}} = \phi^T_k (\Delta A - \lambda_k \Delta W) \phi_k,
\end{equation}
\normalsize
and for eigenvector $\phi_k$,

\small
\begin{equation}
    \frac{\partial \phi_{k}}{\partial p_{ij}} = \mu_{ij} + c\: \phi,
\end{equation}
\normalsize
where $\mu$ is the result of the following equation,
\small
\begin{equation}
   (W-A) \mu_{ij} = F_i,
\end{equation}
\normalsize
and $F_i$ is defined as,
\small
\begin{equation}
F_{i} = \phi^T \left( \frac{\partial W_{k}}{\partial p_{ij}} - \lambda_i \frac{\partial A_{k}}{\partial d_{ij}}\right) \phi A \phi - \left( \frac{\partial W_{k}}{\partial p_{ij}} - \lambda_i \frac{\partial A_{k}}{\partial p_{ij}}\right) \phi
\end{equation}
\normalsize
and $c$ as
\small
\begin{equation}
    c = -\mu A \phi - \frac{1}{2} \phi^T \Delta A \phi.
\end{equation}
\normalsize

With the eigenbasis gradients at hand, we can differentiate any given spectral descriptor, i.e. the heat kernel signature wrt. to any parameter $p$ on a manifold $M$,

\begin{equation}
\label{eq:dhks}
    \frac{\partial h_t}{\partial p_{ij}} = \sum - t e^{-\lambda  t}  \phi^2  \frac{\partial \lambda}{\partial p_{ij}} + 2  e^{-\lambda  t}  \phi \frac{\partial \phi}{\partial p_{ij}}
\end{equation}
which allows us to do an end-to-end analysis, back propagating the loss from the spectral descriptor down to the individual parameter.

\subsection{Learning on mesh edges}
Having established the groundwork, we move on setting up the parameters we want to learn on a manifold and first approach learning the Riemannian metric $g$, which by default is defined on the mesh edge as its length $d_{ij}$. For this we inherit the gradients for the stiffness matrix $W$ from \cite{gahm2018riemannian}, but introduce the gradients for the mass matrix $A$ using Voronoi cells in the following. As the following gradients are similar in nature, we demonstrate them by giving one example and refer the reader to the supplementary materials for detailed gradients and their derivation, which are all derived by basic calculus.

\begin{figure}[H]
\centering
\begin{tikzpicture}


\coordinate (A) at (0.4,0.2); 
\coordinate (B) at (3,0);
\coordinate (C) at (1.5,2);
\coordinate (D) at (3.6,1.8); 

\draw[pattern=north west lines, pattern color=blue!30] (A) -- (B) -- (C) -- cycle;
\draw[pattern=north west lines, pattern color=blue!30] (C) -- (D) -- (B);

\draw (A) -- (B) -- (C) -- cycle;
\draw (B) -- (D) -- (C);
\draw [line width=0.5mm, red] (C) -- (B);

\foreach \point/\label in { D/$k_2$}
    \fill (\point) circle (2pt) node[right] {\label};
\foreach \point/\label in {C/$i$}
    \fill (\point) circle (2pt) node[above] {\label};
\foreach \point/\label in {B/$j$}
    \fill (\point) circle (2pt) node[below] {\label};
\foreach \point/\label in {A/$k_1$}
    \fill (\point) circle (2pt) node[left] {\label};
\pic["$\gamma_1$", draw=black, angle radius=0.7cm] {angle = A--C--B};
\pic["$\gamma_2$", draw=black, angle radius=0.7cm] {angle = B--C--D};
\pic["$\alpha_2$", draw=black, angle radius=0.8cm] {angle = C--D--B};
\pic["$\delta_2$", draw=black, angle radius=0.7cm] {angle = D--B--C};
\pic["$\delta_1$", draw=black, angle radius=0.7cm] {angle = C--B--A};
\pic["$\alpha_1$", draw=black, angle radius=0.8cm] {angle = B--A--C};
\end{tikzpicture}

\caption{Edge flap surrounding edge ij. Adjusting the Riemannian metric, which is highlighted in red, has an impact on all the angles and distances shown and are hence part of the gradients introduced.}
\label{fig:edgeflap}
\end{figure}
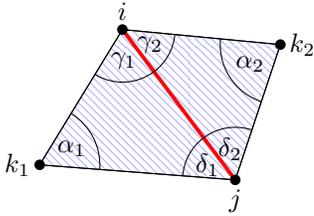

Adjusting any given Riemannian metric has an impact on the surrounding edge flap and thus angles and distances change within this edge flap and the gradients for RiemannNet are derived from the components depicted in Figure \ref{fig:edgeflap}.

We introduce the gradients for the mass matrix $A$ for the module RiemannNet:
\begin{dmath}
      \frac{\partial A_{ii}}{\partial d_{ij}} = \frac{1}{8}  \left(\left(\scriptstyle{-}\dfrac{d_{ij}^3}{\sin^2(\alpha_{1}) 2T_1} + \displaystyle\cot(\alpha_{1}) 2 d_{ij}\right) \scriptstyle{+} \left(\dfrac{d_{jk}^3 \cos(\gamma_{1})}{\sin^2(\delta_{1}) 2T_1}\right)-  \left(\dfrac{d_{ij}^3}{\sin^2(\alpha_{2}) 2T_2} + \displaystyle\cot(\alpha_{2}) 2 d_{ij}\right) \scriptstyle{+} \left(\dfrac{d_{jk}^3 \cos(\gamma_{2})}{\sin^2(\delta_{2}) 2T_2}\right)\right),
\end{dmath}
where $T$ is the area of the triangle.
As triangles have to obey the triangular inequality to be valid, specific weightings can render a triangle invalid, resulting in a failure to solve for the eigenbasis. To mitigate this, \cite{gahm2018riemannian} proposed to use Rosen's gradient projection method \cite{rosen1960gradient}, which projects the gradient into the search direction by taking the triangular inequality into account. As our proposed method takes a different strategy to weight the Riemannian metric, we treat it as a metric nearness problem and solve it in the forward-pass as proposed in \cite{DhillonST04}. This iterative algorithm first weights the triangle with the given weights, which might result in an invalid triangle. Then for every rotation $(ijk, jki, kij)$ of the triangle, the violation is calculated with $d_{ik} + d_{jk} - d_{ij}$ which then is used to derive correction terms with $\frac{1}{3} (e_{ij} - e_{ik} - e_{jk})$. This corrective term is then subtracted from the violating edge term $e_{ij}$ and is added onto the remaining edge terms $(e_{ik}, e_{jk})$. This routine is repeated until no triangular inequality is violated. As collinear vertices in a triangle or triangles very close to having collinear vertices produce exploding gradients, we add a small term $\epsilon$ to the corrective term. This results in a much smoother gradient and prevents invalid eigenbases.


\subsection{Learning on mesh faces}
ALBO was introduced by \cite{andreux2015anisotropic} and adopted in the wider literature in various neural network applications \cite{boscaini2016learning}\cite{boscaini2016anisotropic}\cite{li2020shape} proving the potential beyond the standard LBO. However, the increased power comes at the cost of precalculating ALBO for every combination of anisotropic factor $\alpha$ and angle $\theta$. Furthermore, as both parameters are discrete, important discriminativity might lie between two discrete values and might go unused. To avoid confusion with the notation used in this paper, we label the anisotropic factors as $a_x = \alpha_x$.

As the presented work allows the propagation of a loss function through the LBO eigenbasis, we take a step further and instead of just using ALBO with its standard parameters, we learn them from given data by face-wise perceptrons. By this we can learn a powerful ALBO which is highly adaptive over the manifold, which would otherwise be very hard to obtain with traditional methods.

We adopt the approach from \cite{andreux2015anisotropic} with learning the anisotropic factor $a$ into the principal curvature directions and the extension to the rotation parameter $\theta$ introduced later in \cite{boscaini2016anisotropic}. Additionally, we extend ALBO to learn into both directions at the same time, which we term as ALBO+. Our ablation study shows that there is a significant benefit in allowing the neural network to learn in both directions.

A more general form of equation \ref{eq:LBOFIRST} can be expressed as,
\begin{equation}
   \Delta_D f =  \nabla D(\nabla f)
\end{equation}
where $D$ is referred to as \textit{thermal conductivity} tensor and is defined by a $2\times2$ matrix and controls the direction and anisotropic level of the diffusion.

\begin{figure}[H]
\centering

\begin{tikzpicture}


\coordinate (A) at (0.4,0.2); 
\coordinate (B) at (3,0);
\coordinate (i) at (0,1);
\coordinate (j) at (1.9,2.2);
\coordinate (k) at (1.65,0.74);
\coordinate (C) at (1.5,2);
\coordinate (D) at (3.6,1.8); 

\draw[pattern=north west lines, pattern color=blue!30] (A) -- (B) -- (C) -- cycle;
\draw[pattern=north west lines, pattern color=blue!30] (C) -- (D) -- (B);


\draw (A) -- (B) -- (C) -- cycle;
\draw (B) -- (D) -- (C);
\draw[-{Stealth[black]}] (A) -- ($(A)!0.6!(B)$) node[pos=0.5, below] {$e_{k1j}$};
\draw[-{Stealth[black]}] (A) -- ($(A)!0.6!(C)$) node[pos=0.9, left] {$e_{k1i}$};

\draw[-{Stealth[black]}] (D) -- ($(D)!0.6!(C)$) node[pos=0.5, above] {$e_{k2i}$};
\draw[-{Stealth[black]}] (D) -- ($(D)!0.6!(B)$) node[pos=0.7, right] {$e_{k2j}$};

\foreach \point/\label in { D/$k_2$}
    \fill (\point) circle (2pt) node[right] {\label};
\foreach \point/\label in {C/$i$}
    \fill (\point) circle (2pt) node[above] {\label};
\foreach \point/\label in {B/$j$}
    \fill (\point) circle (2pt) node[below] {\label};
\foreach \point/\label in {A/$k_1$}
    \fill (\point) circle (2pt) node[left] {\label};

\pic["$\alpha_2$", draw=black, angle radius=0.7cm] {angle = C--D--B};

\draw[-{Stealth[red]}, red] (barycentric cs:A=1,B=1,C=1) -- (0,1) node[pos=1, left] {$v_m$};
\draw[-{Stealth[red]}, red] (barycentric cs:A=1,B=1,C=1) -- (1.9,2.2) node[above, above] {$v_M$};
\pic["$\alpha_1$", draw=black, angle radius=0.7cm] {angle = B--A--C};

\pic [draw,red, angle radius=0.3cm, angle eccentricity=.5,pic text=.]{right angle = j--k--i};

\end{tikzpicture}

\caption{Orthonormal reference frame on a face depicted by $v_m$ and $v_M$ based on the principal curvature directions, which is highlighted red. Scaling the directions $v_m$ and $v_M$ by $a$ can change the anisotropy on the given face, which results in scaling dot product, between the corresponding edge and the direction.}
\label{fig:aniso2}
\end{figure}
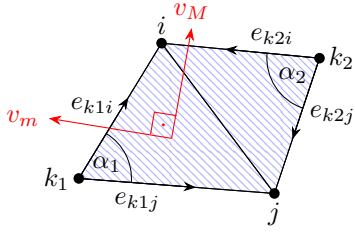

Andreux et al. \cite{andreux2015anisotropic} proposed to use a tensor based on the principal curvature directions ($\kappa_m$, $\kappa_M$) forming an orthonormal basis $V_{ijk} = (v_m, v_M)$ on mesh faces,
\begin{equation}
    D_{a}  = \begin{pmatrix} \psi^m_{a}(\kappa_m, \kappa_M) & 0\\ 0 & \psi^M_{a}(\kappa_m, \kappa_M) \end{pmatrix},
\end{equation}
where $a$ controls the degree of anisotropy as illustrated in red in Figure \ref{fig:aniso2}.

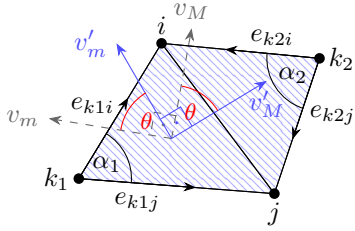
\begin{figure}[H]
\centering
\begin{tikzpicture}


\coordinate (A) at (0.4,0.2); 
\coordinate (B) at (3,0);
\coordinate (i) at (0,1);
\coordinate (j) at (1.9,2.2);
\coordinate (k) at (1.65,0.74);
\coordinate (C) at (1.5,2);
\coordinate (D) at (3.6,1.8); 

\draw[pattern=north west lines, pattern color=blue!30] (A) -- (B) -- (C) -- cycle;
\draw[pattern=north west lines, pattern color=blue!30] (C) -- (D) -- (B);


\draw (A) -- (B) -- (C) -- cycle;
\draw (B) -- (D) -- (C);
\draw[-{Stealth[black]}] (A) -- ($(A)!0.6!(B)$) node[pos=0.5, below] {$e_{k1j}$};
\draw[-{Stealth[black]}] (A) -- ($(A)!0.6!(C)$) node[pos=0.9, left] {$e_{k1i}$};

\draw[-{Stealth[black]}] (D) -- ($(D)!0.6!(C)$) node[pos=0.5, above] {$e_{k2i}$};
\draw[-{Stealth[black]}] (D) -- ($(D)!0.6!(B)$) node[pos=0.7, right] {$e_{k2j}$};

\foreach \point/\label in { D/$k_2$}
    \fill (\point) circle (2pt) node[right] {\label};
\foreach \point/\label in {C/$i$}
    \fill (\point) circle (2pt) node[above] {\label};
\foreach \point/\label in {B/$j$}
    \fill (\point) circle (2pt) node[below] {\label};
\foreach \point/\label in {A/$k_1$}
    \fill (\point) circle (2pt) node[left] {\label};

\pic["$\alpha_2$", draw=black, angle radius=0.7cm] {angle = C--D--B};

\draw[dashed,-{Stealth[black!60]}, black!60] (barycentric cs:A=1,B=1,C=1) -- (0,1) node[pos=1, left] {$v_m$};
\draw[dashed,-{Stealth[black!60]}, black!60] (barycentric cs:A=1,B=1,C=1) -- (1.9,2.2) node[above, above] {$v_M$};
\pic["$\alpha_1$", draw=black, angle radius=0.7cm] {angle = B--A--C};

\pic [dashed, draw,black!60, angle radius=0.3cm, angle eccentricity=.5,pic text=.]{right angle = j--k--i};

\coordinate (i) at (0.9,2);
\coordinate (j) at (2.9,1.5);
\coordinate (k) at (1.65,0.74);

\draw[-{Stealth[blue!70]}, blue!70] (barycentric cs:A=1,B=1,C=1) -- (0.9,2) node[pos=1, left] {$v_m'$};
\draw[-{Stealth[blue!70]}, blue!70] (barycentric cs:A=1,B=1,C=1) -- (2.9,1.5) node[below, below] {$v_M'$};

\pic [draw,blue!70, angle radius=0.3cm, angle eccentricity=.5,pic text=.]{right angle = j--k--i};

\coordinate (i) at (0,1);
\coordinate (j) at (0.9,2);
\coordinate (k) at (1.65,0.74);

\pic [draw,red, angle radius=0.7cm,pic text=\small$\theta$]{angle = j--k--i};

\coordinate (i) at (1.9,2.2);
\coordinate (j) at (2.9,1.5);
\coordinate (k) at (1.65,0.74);

\pic [draw,red, angle radius=0.7cm,pic text=\small$\theta$]{angle = j--k--i};

\end{tikzpicture}

\caption{Orthonormal reference frame on a face depicted by $v_m$ and $v_M$ based on the principal curvature directions, which is highlighted blue. It is rotated by $\theta$, which is depicted red. Besides, scaling the directions $v_m$ and $v_M$ rotating the directions has a direct impact on the used dot products.}
\label{fig:aniso1}
\end{figure}

Later \cite{boscaini2016anisotropic} proposed to include the rotation parameter $\theta$ by rotating $D_{a}$,
\begin{equation}
    D_{a\theta} = R_{\theta}D_{a}(x)R_{\theta}^{T},
\end{equation}
where $R_\theta$ denotes the rotation matrix as illustrated red in Figure \ref{fig:aniso1}.

While the authors \cite{boscaini2016anisotropic} fix the anisotropic factor in one direction and varying the other direction, we apply ALBO+ in all our experiments as it is more powerful then the standard ALBO. 

This results in a change in the stiffness matrix W, which now is defined as:
\begin{equation}
\label{eq:albostiffness}
W_D(i,j) = \begin{cases}
    \frac{\langle e_{k1j}, e_{k1i} \rangle_H}{\sin \alpha_{1}} +  \frac{\langle e_{k2j}, e_{k2i} \rangle_H}{\sin \alpha_{ 2 }} , & \text{if }(i,j) \in E\\
    - \sum_{k \neq i} W(i,k), & \text{if }i=j\\
0, & \text{otherwise},
\end{cases}  
\end{equation}
where   
\begin{equation}
    \langle e_{kj}, e_{ki} \rangle_H = e_{kj}^T \underbrace{V_{ijk}D_{\alpha\theta}V_{ijk}^T}_{H}e_{ki}
\end{equation}
and $e$ being the respective edge on the triangle depicted in Figure \ref{fig:aniso1}.

As the changes of equation \ref{eq:albostiffness} impact only the stiffness matrix $W$, we adjust Nelson's method given in equation \ref{eq:generalizedNelson} by omitting the gradients for the mass matrix $A$,
\begin{equation}
    \frac{\partial \lambda_{k}}{\partial p_{ij}} = \phi^T_k (-\lambda_k \Delta W) \phi_k,
\end{equation}
\begin{equation}
F_{i} = \phi^T \left( \frac{\partial W_{k}}{\partial p_{ij}}\right) \phi A \phi - \left( \frac{\partial W_{k}}{\partial p_{ij}}\right) \phi
\end{equation}

\subsubsection{Gradient sign of dot products}
The gradient of the dot product in ALBO, i.e. edge $e_{k1}$ and $v_m$ changes its sign, depending on which side $e_{k1}$ is wrt. $v_m$. To calculate the sign, we use the product of the cross products,

\begin{equation}
    sign(x) = e_{k1i} \times e_{k1j} *  e_{k1j} \times v_{m}
\end{equation}
This sign is then applied to all dot products in the ALBO to correct the sign.

\subsubsection{Anisotropic factors $a_1$,  $a_2$}
\begin{equation}
    \frac{\partial W_{ii}}{\partial a_{xf}} = \frac{1}{2} \langle e_{ki} v_{M} \rangle^2  (\cot(\alpha_1) + \cot(\alpha_2)),
\end{equation}
where $x \in \{1,2\}$ is the corresponding factor scaling each principal curvature direction in Figure \ref{fig:aniso2} and $f$ any given face on the mesh, e.g. $ijk_1$ in Figure \ref{fig:aniso2}.

\subsubsection{Rotation parameter $\theta$}
As the rotation only occurs on the orthogonal plane formed by the principal directions (see Figure \ref{fig:aniso1}), the problem is reduced onto the angle only and is given by:
\begin{dmath}
    \frac{\partial W_{ii}}{\partial \theta_{f}} = -(\cot(\alpha_1) + \cot(\alpha_2)) (a_1 \sin{\cos^{-1}{\langle e_{ki} v_{M} \rangle }} {\langle e_{kj} v_{M} \rangle } + a_2 \sin{\cos^{-1}{\langle e_{ki} v_{m} \rangle }} {\langle e_{kj} v_{m} \rangle }).
\end{dmath}

\subsubsection{ALBONet and RiemannNet}
As the previously introduced RiemannNet gradients assume the LBO to be isotropic, the introduction of ALBONet requires a new set of RiemannNet gradients for the stiffness matrix for the anisotropic case,\\
\begin{dmath}
          \dfrac{\partial W_{ii}}{\partial d_{ij}} =  - \frac{1}{2} (a_{1}  \langle e_{ki}v_{M} \rangle ^2 + a_{2} \langle e_{ki}v_{m} \rangle^2 ) \frac{d_{ij}}{2T}  \frac{1}{sin(\delta)^2} - \frac{(d_{ik}  \cos(\gamma)  )}{2T}  \frac{1}{\sin(\alpha)^2} + (a_{1}  2  \sin \cos^{-1}(\langle e_{ki}v_{M} \rangle)) \langle e_{ki}v_{M} \rangle   \frac{d_{ik}  \cos(\gamma)}{2T} + a_{2}  2  \sin \cos^{-1}(\langle e_{ki}v_{m} \rangle)) \langle e_{ki}v_{m} \rangle   \frac{d_{ik}  \cos(\gamma)}{2T})  (\cot(\alpha_1) + \cot(\alpha_2))
\end{dmath}

\subsection{Learning on mesh vertices}
Weighting the Voronoi cell has been proven to be beneficial by \cite{shi2011conformal}\cite{limberger2018curvature}. The challenge however remains in the strategy of weighting the Voronoi cell. Similar to \cite{shi2011conformal}, we propose to backpropagate the loss through the LBO eigenbasis but learn the Voronoi cell weighting $A_{ii}$ by vertex-wise perceptrons instead.

The changes for VoronoiNet are limited to the mass matrix $A$. Therefore we adjust Nelson's method given in equation \ref{eq:generalizedNelson} by omitting the gradients for the stiffness matrix $W$,
\begin{equation}
    \frac{\partial \lambda_{k}}{\partial m_{ij}} = \phi^T_k (\Delta A) \phi_k,
\end{equation}
\begin{equation}
F_{i} = \phi^T \left(- \lambda_i \frac{\partial A_{k}}{\partial d_{ij}}\right) \phi A \phi - \left( \lambda_i \frac{\partial A_{k}}{\partial d_{ij}}\right) \phi
\end{equation}

and introduce the gradients for VoronoiNet as:
\begin{equation}
    \dfrac{\partial A_{ii}}{\partial v_{ii}} = A_{ii}.
\end{equation}

\subsection{LBONet Architecture}
Our contribution can be plugged into any recent work which we refer to as "backend" (see Figure \ref{fig:arch}), however we demonstrate in our ablation study that even with a traditional backend (a single shared MLP), LBONet is able to outperform neural networks with comprehensive backends such as \cite{wang2019dynamic} \cite{hanocka2019meshcnn} \cite{qi2017pointnet++}. The performance comparison to \cite{qi2017pointnet++} is particularly interesting because our ablation proves that the performance gain obtained by PointNet++, can be similarly obtained by learning with LBONet through the LBO eigenbasis. The upside this work presents however is, that it can still be combined with a more recent backend to gain an even better performance. LBONet outperforms on many benchmarks or achieves competitive results. We show this with one backend derived from PointNet \cite{qi2017pointnet} for retrieval and classification. The study of pooling operations in \cite{yu2020second} reveals that average pooling is clearly outperforming max pooling; Therefore, we apply average pooling rather than max pooling. For segmentation tasks we derive the backend from point transformer \cite{wu2022point}, by feeding the resulting spectral descriptor into the transformer and into the dimensionality reduction layer by a skip connection. Finally, for correspondence tasks we observed that the setup of DiffusionNet \cite{sharp2022diffusionnet} is a great fit as it is making use of heat kernel signature and functional maps, both elements which our method is learning. LBONet can be combined with any backend as long as the gradients backpropagated to the spectral descriptors are useful (we observed with deep networks vanishing gradients, which we circumvented with skip connections). Furthermore to speed up learning, we implemented a routine to learn LBONet directly on the given data and then combined it with the backend for fine tuning. While it is possible to run the experiments in a single go, this approach proved to be more efficient with comparable results.

Furthermore to stabilize and increase the speed of learning features with LBONet, we employ batch normalization, after every shared MLP.

\subsection{Implementation}
\paragraph{Forward pass}
We solve the generalized eigenvalue problem on the CPU using SciPy's \cite{Virtanen2020} implementation of restarted Arnoldi methods \cite{lehoucq1998arpack}. The neural networks are implemented in PyTorch \cite{paszke2019pytorch} and are trained on a single GPU except for the correspondence task, which due its mesh resolution required to be trained on an Nvidia H100.
\paragraph{Backward pass}
Standard libraries such as PyTorch or Tensorflow only allow to differentiate the full spectrum of a matrix, creating two problems: firstly the sparsity of the data is not respected, and secondly, as the most important information is contained within the lower frequencies, one cannot target specific eigenfrequencies. HodgeNet \cite{smirnov2021hodgenet} takes an alternative approach to mitigate these problems. As of writing this paper, the \textit{LOBPCG} method in PyTorch does not support backpropagation when both stiffness and mass matrix are provided, moreover it cannot deal with sparse matrices.

\begin{table}[H]
\centering
\begin{tabular}{@{}lllll@{}}

Vertices & 1024 & 2048 & 4096 &  \\ \midrule
RiemannNet & 0.09 & 0.51 & 3.4  &  \\
ALBONet    &   0.05   &   0.4   &  2.5    &  \\
ALBONet+    &   0.08   &   0.6   &  3.3    &  \\
VoronoiNet &  0.03    &   0.2   &  0.8    &  \\ 
\midrule
\end{tabular}
\caption{Backpropagation runtimes on varying mesh resolution on a RTX 4090 GPU given in seconds.}\label{wrap-tab:1}
\end{table}

LBONet introduces a framework to tackle both these problems by respecting the sparsity of the data and by addressing only a certain range of frequencies (first eigenfrequencies can be skipped for some tasks) while differentiating the eigenbasis. We introduce custom functions to deal with sparse data and a subset of the eigenfrequencies. Despite the computational complexity, the presented routine is fully vectorized across both dimensions—edges/faces/vertices and frequencies - enabling the calculation of the entire mesh gradient for the full eigenbasis simultaneously. This results in computations completed within a reasonable timeframe. Refer to Table \ref{wrap-tab:1} for the running times of the presented modules. In contrast to HodgeNet, the backpropagation procedure of LBONet does not require the computation of additional eigenfrequencies. Conversely, we observed the learning effect to be extending to frequencies, which were not used during the training process. This effect could allude that the features learnt by LBONet could generalize across frequencies as well, which is subject to further research. As the eigenbasis gradient can be very sensitive and take extreme values in certain cases we apply gradient clipping, to mediate numerical instabilities during the training process. Although LBONet is computationally expensive to train, it has a relatively small number of parameters, as all neural network components are shared and lightweight. The features it learns are intrinsic to the manifold, enabling improvements over the standard LBO to emerge after only a few training epochs, without the need for extensive training. In its default configuration, LBONet contains approximately 41,000 parameters - a negligible size compared to other frameworks. Thanks to the network’s compactness and the fundamental nature of the learned features, LBONet generalizes well to unseen data, as demonstrated by our evaluation. We release our code for reproducibility and to facilitate further research at: \url{https://github.com/yioguz/LBONet}.

\section{Evaluation}
We quantitatively evaluate LBONet across several tasks, including retrieval and classification. While retrieval and classification focus on learning global descriptors, we further demonstrate the effectiveness of LBONet in segmentation and correspondence tasks, which require descriptors with high local fidelity. For each benchmark, we also report baseline results obtained by disabling LBONet, while keeping the rest of the network architecture including its capacity unchanged. This setup ensures that any observed performance gains can be attributed specifically to the inclusion of LBONet, rather than to differences in model capacity or design.
Furthermore, to highlight the net contribution of our method, we indicate the performance of the standard LBO, which serves as our baseline against LBONet in detail. In all experiments, LBONet is trained using the first 32 eigenfrequencies. We employ EdgeConv with a neighborhood size of 20, along with two layers of shared MLPs (each with 32 neurons), using LeakyReLU activation and batch normalization.
\label{chap:experiments}

\subsection{Retrieval}

We evaluate our model on several established benchmarks: SHREC'11 \cite{lian2011shape}, SHREC'14 \cite{pickup2014shrec}, SHREC'15 \cite{10.5555/2852282.2852307}, and ShapeNetCore55 \cite{yi2016scalable}.
The SHREC'11 benchmark comprises 600 manifolds divided into 30 classes, with 20 shapes per class. Following the protocol in \cite{sharp2022diffusionnet}, we train on only 10 shapes per class. To ensure robustness, we repeat the experiment five times using different random training subsets and report the average performance.
The SHREC'14 benchmark consists of two datasets: one based on real scans and the other on synthetic manifolds. The real benchmark is particularly challenging, as many algorithms struggle to achieve high accuracy on it.
For SHREC'15, we follow the protocol of \cite{qi2017pointnet++} and use k-fold cross-validation to compute the performance metrics.
Lastly, ShapeNetCore55 contains over 51,000 manifolds spanning 55 categories, providing a large-scale benchmark for evaluating generalization across diverse shapes.

We report our retrieval results using the standard evaluation metrics introduced by \cite{shilane2004princeton}, which capture different aspects of retrieval quality, including Nearest Neighbor (NN), First Tier (FT), Second Tier (ST), and Discounted Cumulative Gain (DCG). For ShapeNet, we report mean Average Precision (mAP), as it has become the standard metric for this dataset and facilitates comparison with a broader range of existing methods in the literature.

We achieve competitive results on SHREC'11 (Table \ref{tab:res11}), reaching perfect performance. On both SHREC'14 benchmarks (Tables \ref{tab:res14} and \ref{tab:res14s}), we outperform all other methods, with particularly striking results on the ‘real’ benchmark. The significant improvement in the First Tier metric demonstrates LBONet’s capability to fully exploit the potential of the LBO. Since the SHREC'11 dataset is a subset of SHREC'15, we observe a similar performance gain on SHREC'15 (Table \ref{tab:res15}), achieving near-perfect results. Notably, the inclusion of manifolds with noise and topological variations has minimal impact on performance. Our results on ShapeNetCore55 are reported in Table \ref{tab:shapenet55ret}; while we do not outperform all methods, we achieve a substantial improvement over the baseline and maintain competitive performance compared to multi-view CNN approaches.

\label{sec:evaluation}
\begin{table}[H]
\begin{center}
\begin{tabular}{ c c }

Method & mAP \\
\midrule
RotNet\cite{kanezaki2018rotationnet} & 77.2 \\
ViewGCN\cite{wei2020view} & 78.4 \\
MVTN \cite{hamdi2021mvtn} & \textbf{82.9} \\ 
Baseline & 75.3 \\ 
\textbf{LBONet} & 79.1  \\ 
\midrule
\end{tabular}
\caption{\label{tab:shapenet55ret}Retrieval Results on ShapeNetCore55 (in \%)}
\end{center}
\end{table}
\begin{table}[H]
\begin{center}
\setlength{\tabcolsep}{2pt}
\begin{tabular}{ c c c c c}
\setlength{\tabcolsep}{2pt}

Method & NN & FT & ST & DCG\\
\hline
SD-GDM-meshSIFT \cite{lian2011shape} & \textbf{100} & 97.20 & 99.01 & 99.55\\ 
FV-IWKS\cite{limberger2015feature} & 99.83 & 95.91 & 98.60 & 99.37\\
KLBO \cite{limberger2017spectral}  & \textbf{100} & \textbf{100} & \textbf{100} & \textbf{100}\\ 
Baseline & 99.67 & 98.04 & 98.30 & 99.11\\
\textbf{LBONet} & \textbf{100} & \textbf{100} & \textbf{100} & \textbf{100}\\ 
\midrule
\end{tabular}
\caption{\label{tab:res11}Retrieval Results on SHREC'11 Dataset (in \%)}
\end{center}
\end{table}

\begin{table}[H]
\begin{center}
\setlength{\tabcolsep}{2pt}
\begin{tabular}{ c c c c c c}
Method & NN & FT & ST & DCG\\
\hline
supDLtrain \cite{litman2014supervised} & 79.3 & 72.7 & 91.4 & 89.1\\ 
DeepGM\cite{luciano2019global} & 72.5 & 53.6 & 82.7 & 78.2\\ 
SOST\cite{yu2020second} & 85.3 & 63.2 & 85.2 & 85.1\\ 
Baseline & 85 & 64.81 & 86.03 & 86.41\\ 
\textbf{LBONet} & \textbf{86.5} & \textbf{77.33} & \textbf{91.14} & \textbf{90.69}\\ 
\midrule
\end{tabular}
\caption{\label{tab:res14}Retrieval Results on SHREC'14 'real' Dataset (in \%)}
\end{center}
\end{table}

\begin{table}[H]
\begin{center}
\setlength{\tabcolsep}{2pt}
\begin{tabular}{ c c c c c c}
\setlength{\tabcolsep}{0pt}
Method & NN & FT & ST & DCG\\
\midrule
supDLtrain \cite{litman2014supervised} & 96.0 & 88.7 & \textbf{99.1} & 97.5\\ 
DeepGM \cite{luciano2019global} & 99.3 & 81.4 & 98.3  & 96.7\\ 
3D-DL\cite{bu2014learning} & 92.3 & 76.0 & 91.1 & 92 .1\\
KLBO \cite{limberger2017spectral}  & 88.83 & 81.58 & 95.12 & 94.03\\ 
Baseline & 97.33 & 81.40 & 95.25 & 95.30\\ 
\textbf{LBONet} & \textbf{99.33} & \textbf{89.21} & 97.95 & \textbf{97.87}\\ 
\midrule
\end{tabular}
\caption{\label{tab:res14s}Retrieval Results on SHREC'14 'synthetic' Dataset (in \%)}
\end{center}
\end{table}

\begin{table}[H]
\begin{center}
\setlength{\tabcolsep}{2pt}
\begin{tabular}{ c c c c c }
\setlength{\tabcolsep}{2pt}
Method & NN & FT & ST & DCG\\
\midrule
HAPT \cite{giachetti2012radial} & 99.8 & 96.6 & 98.2 & 99.2\\ 
DeepGM \cite{luciano2019global} & 99.3 & 94.0 & 97.2 & 98.8\\ 
AP-SOST-Net\cite{yu2020second} & 99.1 & 93.7 & 96.5 & 97.5\\
KLBO \cite{limberger2017spectral}  & \textbf{99.92} & 98.59 & 99.33 & \textbf{99.59}\\ 
Baseline & 97.2 & 96.4 & 98.5 & 98.48\\
\textbf{LBONet} & 99.20 & \textbf{98.80} & \textbf{99.70} & 99.48\\ 
\midrule
\end{tabular}
\caption{\label{tab:res15}Retrieval Results on SHREC'15 Dataset (in \%)}
\end{center}
\end{table}

\subsection{Classification}
In order to compare the presented work to the broader literature, we conduct the experiments on SHREC'11 and SHREC'15 in a classification setting. Therefore we employ the same settings as in the retrieval case, with the exception of using cross entropy loss for learning.

\begin{figure}[h]
  
\includegraphics[scale=0.13,trim=12cm 5cm 0cm 8cm]{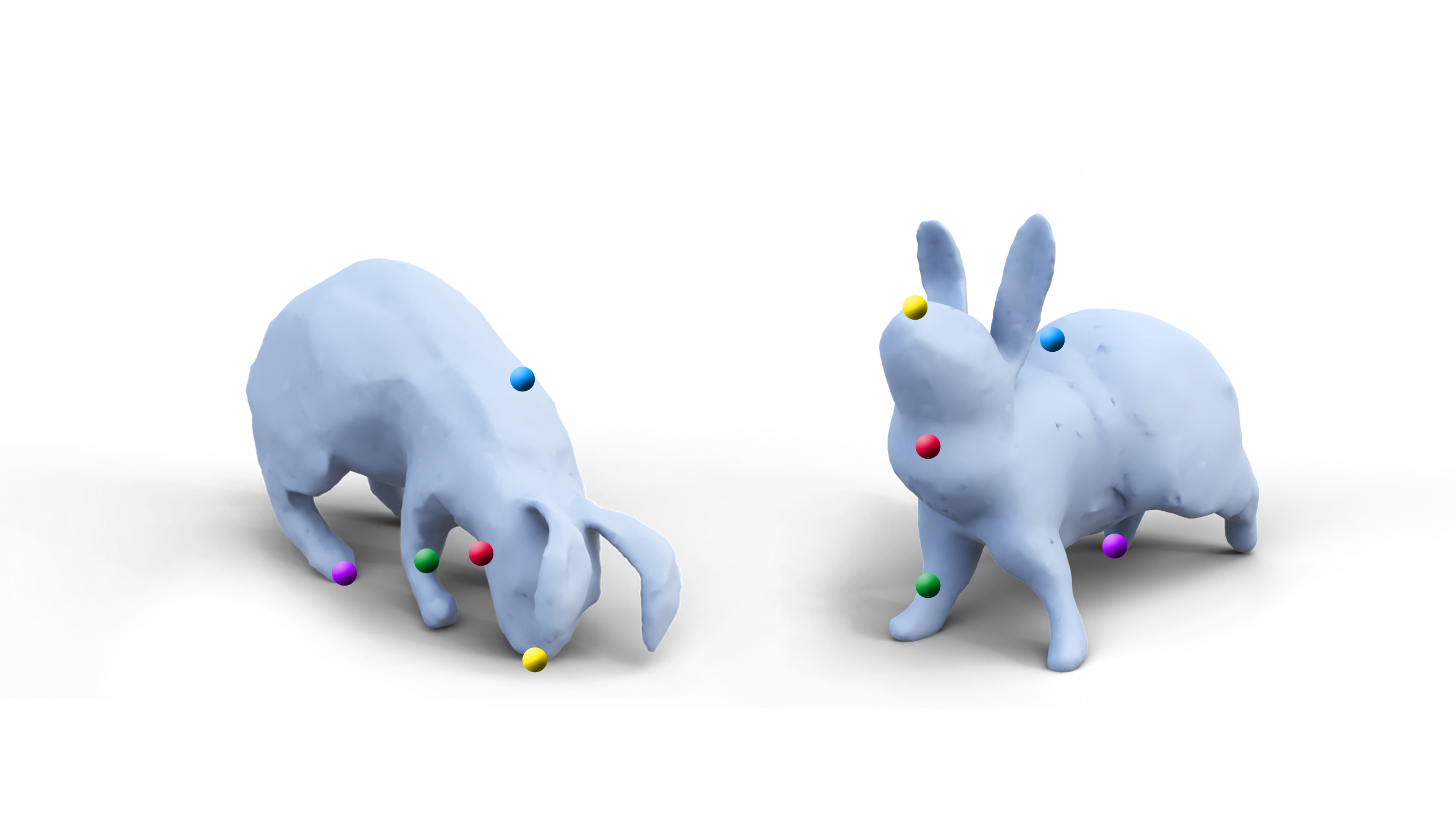}
\begin{tikzpicture}
\definecolor{zellow}{RGB}{255,239,62}
\definecolor{ylue}{RGB}{0,142,242}
\definecolor{gleen}{RGB}{80,181,111}
\definecolor{mag}{RGB}{196,70,255}
\definecolor{zed}{RGB}{229,44,92}
        \begin{axis}[
            ybar,
            symbolic x coords={A, B, C, D, E},
            xtick=data,
            xticklabels={\textcolor{ylue}{\Large\textbullet}, \textcolor{gleen}{\Large\textbullet}, \textcolor{zed}{\Large\textbullet}, \textcolor{zellow}{\Large\textbullet}, \textcolor{mag}{\Large\textbullet}},
            bar width=8pt, 
            enlarge x limits=0.15,
            ymin=0,
            ymax=2.6,
            axis y line*=left,
            axis x line*=bottom,
            ylabel={\(\Sigma^{15}_{t=0}\Delta hks(t)\)},
            xlabel={},
            width=\columnwidth,
            height=0.5\columnwidth,
            legend style={at={(0.5,-0.2)},
                anchor=north,legend columns=-1},
            x tick label style={rotate=0, anchor=north},
            bar width=8pt
        ]

\addlegendimage{area legend, pattern=north east lines, pattern color=gray}
\addlegendentry{LBO}
 \addlegendimage{area legend, fill=gray}
\addlegendentry{LBONet}
            \addplot[%
                ybar,
                pattern=north east lines,
                pattern color=gray,
                bar shift=-4pt, 
                point meta=y
            ] coordinates {(A, 1.5352008) (B, 0.79212505) (C, 2.5609725) (D, 1.3373115) (E, 0.5514306)};
            \addplot[%
                ybar,
                fill=gray,
                bar shift=4pt, 
                point meta=y
            ] coordinates {(A, 1.3431687) (B, 0.55827534) (C, 2.4022477) (D, 1.1602216) (E, 0.40409607)};
            

        \end{axis}
    \end{tikzpicture}
\caption{Heat kernel signature at selected points evaluated with the standard LBO and with the LBONet trained on the SHREC'11 benchmark. Values are given as the difference of the heat kernel at the highlighted points.}
\label{fig:rabbit}
\end{figure}
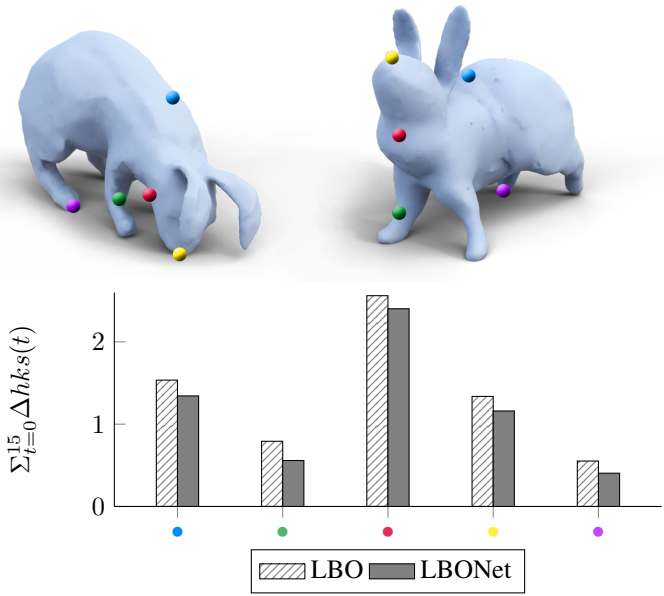

We observe a similar performance gain as in the retrieval case for both benchmarks. The accuracy for SHREC'11 (Table \ref{tab:res151}) and on SHREC'15 (Table \ref{tab:res155}) is perfect. We observed that after applying LBONet that near-isometries are alleviated to some extent as demonstrated in Figure \ref{fig:rabbit}.
\begin{table}[!th]
\begin{center}
\setlength{\tabcolsep}{2pt}
\begin{tabular}{ c c }

Method & Accuracy \\
\midrule
MeshCNN\cite{hanocka2019meshcnn} & 91.0 \\ 
HodgeNet\cite{smirnov2021hodgenet} & 94.67 \\ 
MeshWalker\cite{lahav2020meshwalker} & 97.1 \\ 
PD-MeshNet\cite{milano2020primal} & 99.1 \\
FC\cite{mitchel2021field} & 99.2 \\
DiffusionNet\cite{sharp2022diffusionnet} & 99.7 \\ 
Baseline & 98.0 \\ 
\textbf{LBONet} & \textbf{100.0}  \\ 
\midrule
\end{tabular}

\caption{\label{tab:res151}Classification Results on SHREC'11 Dataset (in \%)}
    
\end{center}
\end{table}
\begin{table*}[th!]
    \small
    \centering
    \begin{tabular}[width=\linewidth]{l|c|p{0.5cm}p{0.4cm}p{0.4cm}p{0.4cm}p{0.5cm}p{0.6cm}p{0.5cm}p{0.5cm}p{0.5cm}p{0.6cm}p{0.6cm}p{0.3cm}p{0.5cm}p{0.6cm}p{0.6cm}p{0.6cm}}
    \hline
    ~        & mean & aero & bag & cap & car & chair & ear & guitar & knife & lamp & laptop & motor & mug & pistol & rocket & skate & table \\ 
    &   & &  &  &  &  & phone &  &   &  &  & &    &    &    & board &  \\ \hline
    \# shapes & & 2690 & 76 & 55 & 898 & 3758 & 69 & 787 & 392 & 1547 & 451 & 202 & 184 & 283 & 66 & 152 & 5271 \\ \hline
    AGCN~\cite{kim2021agcn} &  87.9  & 87.6  & \textbf{92.3} & \textbf{87.3} & \textbf{82.4} & \textbf{92.8} & \textbf{82.3} & \textbf{93.4} & \textbf{89.1} &  \textbf{86.2}  & \textbf{96.5}  & \textbf{80.4} &  \textbf{97.9}  &  86.2   &  69.0 & \textbf{80.0} &  84.4 \\
    SpecCNN~\cite{yi2017syncspeccnn} & 84.8 & 81.6 & 81.7 & 81.9 & 75.2 & 90.2 & 74.9 & 93.0 & 86.1 & 84.7 & 95.6 & 66.7 & 92.7 & 81.6 & 60.6 & 82.9 & 82.1 \\ \hline
    Baseline & 84.7 & 86.1 & 88.9 & 80.3 & 65.0 & 87.7 & 61.5 & 89.3 & 87.6 & 82.4 & 95.4 & 59.9 & 95.5 & 85.4 & 61.6 & 75.7 & 85.0 \\ 
    \textbf{LBONet} & \textbf{89.4} & \textbf{89.1} & 92.0 & 85.5 & 80.9 & 91.0 & 79.8 & 92.4 & 88.7 & 85.5 & 95.4 & 73.7 & 96.6 & \textbf{86.7} & \textbf{78.5} & 77.9 & \textbf{91.1} \\ \hline
    \end{tabular}
    \caption{\textbf{Segmentation results on ShapeNet part dataset.} We report results using the mean Intersection over Union (mIoU, \%) computed on points. Our method is compared against another spectral approach that leverages the LBO \cite{yi2017syncspeccnn}, the competitive AGCN method \cite{kim2021agcn}, as well as our own baseline. LBONet achieves competitive performance in terms of mIoU, demonstrating its effectiveness in the segmentation task.}

    \label{tab:shapenetseg}
\end{table*}

\begin{table}[!th]
\begin{center}
\setlength{\tabcolsep}{2pt}
\begin{tabular}{ c c }

Method & Accuracy \\
\midrule
DeepGM\cite{luciano2019global} & 93.03 \\ 
SpiderCNN\cite{xu2018spidercnn} & 95.8 \\
PointNet++\cite{qi2017pointnet++} & 96.09 \\ 
Baseline & 97.8 \\ 
\textbf{LBONet} & \textbf{100.0}  \\ 
\midrule
\end{tabular}

\caption{\label{tab:res155}Classification Results on SHREC'15 Dataset (in \%)}
    
\end{center}
\end{table}

\subsection{Segmentation}
\begin{figure}
    \centering

        \begin{overpic}[scale=0.25,trim=0cm 0cm 0cm 0cm,clip]{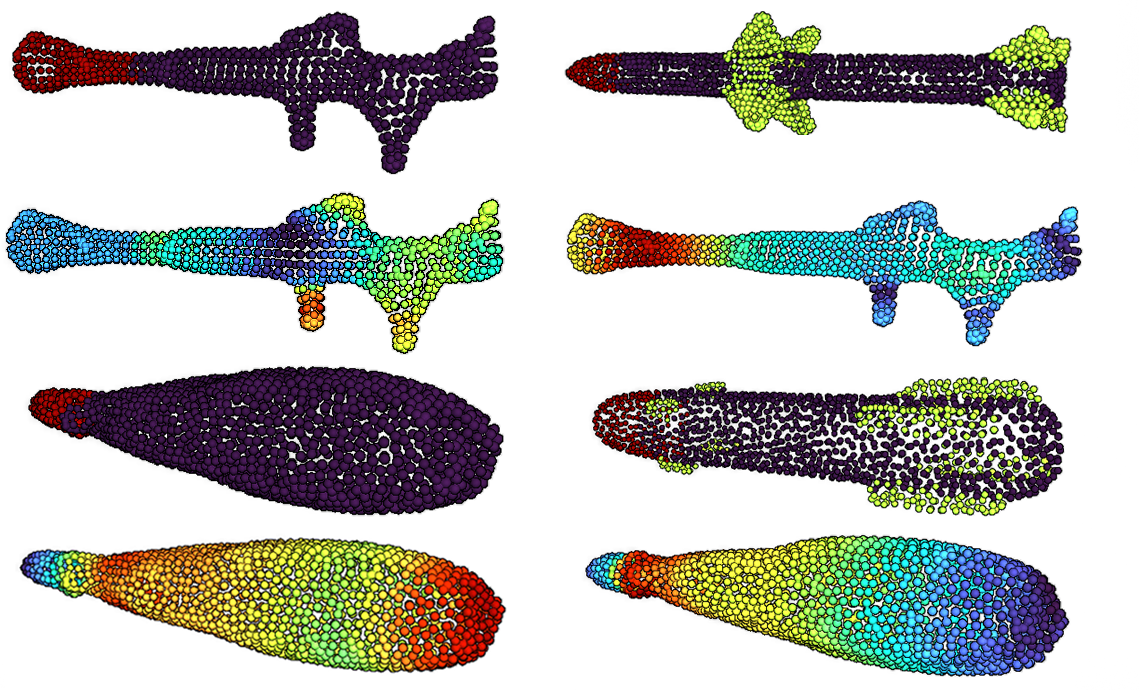}
        \put(1,5){\textbf{g)}}  
        \put(51,5){\textbf{h)}} 
        \put(1,60){\textbf{a)}}  
        \put(51,60){\textbf{b)}} 
        \put(1,45){\textbf{c)}}  
        \put(51,45){\textbf{d)}} 
        \put(1,30){\textbf{e)}}  
        \put(51,30){\textbf{f)}} 
    \end{overpic}
    \caption{LBONet leverages intrinsic and semi-intrinsic features, enabling it to learn orientation- and position-invariant patterns and reducing erroneous segmentations caused by spatial bias. (a) An RPG example where the handle and scope, correctly part of the body, are confused with rocket blades by coordinate-based methods. (b) A rocket instance where reliance on spatial coordinates leads to false feature attribution. (c) Distance from LBONet's HKS descriptors to the centroid of the "blades" region, compared to the baseline. (d) Distance to the main body of the manifold. (e–h) Corresponding plots on a second manifold, confirming the consistency of LBONet’s intrinsic representations.}
            \label{fig:shapenetrockets}
\end{figure}

For shape segmentation, we utilize the ShapeNet Part benchmark introduced by \cite{yi2016scalable}, which comprises 16 object categories, each annotated with 2 to 6 part labels. We adopt the official train/test split for performance evaluation. To ensure compatibility with our pipeline, we preprocess the shapes using the method from \cite{huang2018robust} to obtain watertight meshes. This preprocessing step enables LBONet to operate effectively on the dataset. We present our results in Table \ref{tab:shapenetseg}. Several observations stand out across the categories. LBONet, by virtue of its reliance on intrinsic and semi-intrinsic features, effectively avoids overfitting to spatial cues (a common issue among coordinate-based approaches). Even the state-of-the-art method proposed in \cite{kim2021agcn} is affected by this limitation. For example, in the rocket category, the training set includes a flipped example, which leads coordinate-based methods to misinterpret structural features. Our analysis indicates that such ambiguities can only be robustly resolved by relying on purely intrinsic representations. In Figure \ref{fig:shapenetrockets}, we illustrate the effect of intrinsic versus extrinsic learning. In plot (g), the distance to the blade segment increases in spectral space, most notably around the neck and tail of the rocket, although the spectrum is not explicitly aware of these regions. A similar effect is observed in plot (c). This behavior effectively guides downstream components to avoid hallucinating incorrect segment boundaries. Plots (d) and (h) show that the distance to the body segment is reduced, demonstrating that, in addition to improving robustness against hallucinations, LBONet also contributes to the refinement of actual segmentation boundaries. All features were predicted in the validation set, demonstrating the generalizability of LBONet and its ability to effectively transfer spectral learning from the training data. We observe a significant performance gain over both the baseline and SyncSpecCNN \cite{yi2017syncspeccnn}, and outperform AGCN \cite{kim2021agcn} in several categories as well as in the mean IoU.

The human segmentation benchmark \cite{maron2017convolutional} has been widely used in the literature and features human shapes in different poses, which we will refer to as humans in the following. The COSEG benchmarks \cite{wang2012active} feature vases, chairs, and aliens, which we indicate below by their category. We evaluated on the human test set and for the COSEG benchmarks, we adapt the training setup as in \cite{hanocka2019meshcnn}\cite{milano2020primal}\cite{smirnov2021hodgenet} and split the data set randomly in 85\% for training and 15\% for testing. We report our results based on hard labels in Table \ref{tab:humansSeg}.
\begin{table}[!th]
\begin{center}
\setlength{\tabcolsep}{2pt}
\begin{tabular}{ c c c c c} 
Method & Humans  & Vases & Chairs & Aliens \\
\midrule
MeshCNN\cite{hanocka2019meshcnn} & 85.4 & 92.36 & 92.99 & 96.26\\ 
HodgeNet\cite{smirnov2021hodgenet} & 85.0  & 90.30 & 95.68 & 96.03\\ 
PD-MeshNet\cite{milano2020primal} & 85.6 & 95.36 & 97.23& 98.18 \\
DGCNN \cite{wang2019dynamic} & 89.7 & - & - &  -\\
DiffusionNet \cite{sharp2022diffusionnet} & 91.7 & - &  - & -\\
MeshWalker \cite{lahav2020meshwalker} & 92.7 & - &  -& - \\
FC \cite{mitchel2021field} & \textbf{92.9} & - &  -& -\\
Baseline  & 91.1 & 94.7 &  96.5 & 97.2 \\
\textbf{LBONet} & 92.7  & \textbf{95.6} & \textbf{97.8} & \textbf{98.3}\\ 
\midrule
\end{tabular}

\caption{\label{tab:humansSeg}Segmentation accuracy on humans and COSEG benchmarks (in \%)}

\end{center}
\end{table}
 We outperform most of the methods, including DiffusionNet and MeshCNN. All other methods were already outperformed by the simple architecture used in our ablation study (Table \ref{tab:ablationtable}) and when LBONet is combined with a recent backend such as a point transformer \cite{wu2022point}, it becomes even more powerful as the learning can be based on a highly optimized spectral descriptor provided by LBONet. For a fair comparison we also include the performance of our baseline method \cite{wu2022point} based on the standard heat kernel signature and the net gain of using LBONet becomes apparent. We discuss the effects of LBONet to the LBO eigenbasis in the segmentation tasks in the ablation study, which further reveals how the segments of the manifolds become “encoded” into the LBO eigenbasis, making it easier for descriptors to separate the segments more easily.

\subsection{Functional Correspondence}
For functional correspondence tasks we adapt the supervised setup from \cite{sharp2022diffusionnet} for the FAUST \cite{bogo2014faust} dataset. The FAUST dataset includes both intra- and interclass manifolds of varying degree of near-isometric deformation for training and testing. It serves as a great way to compare the performance gain with the wider literature. We randomly generate all combinations of the 80 manifolds for training and test on the remaining combinations of 20 manifolds. The errors are averaged over all manifolds. We learn the eigenbasis through the heat kernel signature directly using triplet loss using corresponding points and generating dissimilar points on the fly.  We feed the learned heat kernel signature and its eigenbasis into DiffusionNet and achieve a significant reduction in geodesic mean error, proving two facts, first that the learned heat kernel signature is more descriptive than the standard used by DiffusionNet and second, that the eigenbasis is better for constructing functional maps after the learning.

\begin{table}[!th]
\begin{center}
\setlength{\tabcolsep}{2pt}
\begin{tabular}{ ccccc}

\midrule
KPConv \cite{thomas2019kpconv} & 2.9 \\
HSN \cite{Wiersma2020}& 3.3  \\ 
ACSCNN \cite{rampini2019correspondence}&  2.7 \\ 
DiffusionNet\textsuperscript{\textdagger} \cite{sharp2022diffusionnet} & 2.5  \\
Unsupervised \cite{cao2023unsupervised} & 1.5  \\
\textbf{LBONet} & \textbf{1.4}\\
\midrule
\end{tabular}

\caption{\label{tab:res14b} Results on the FAUST \cite{bogo2014faust} benchmark. Values are given in geodesic mean error x 100. Result marked with \textdagger  serves as the baseline method.}
\end{center}
\end{table}
\begin{figure*}[!th]
    \centering

    \includegraphics[scale=0.3,trim=10cm 2cm 8cm 2cm]{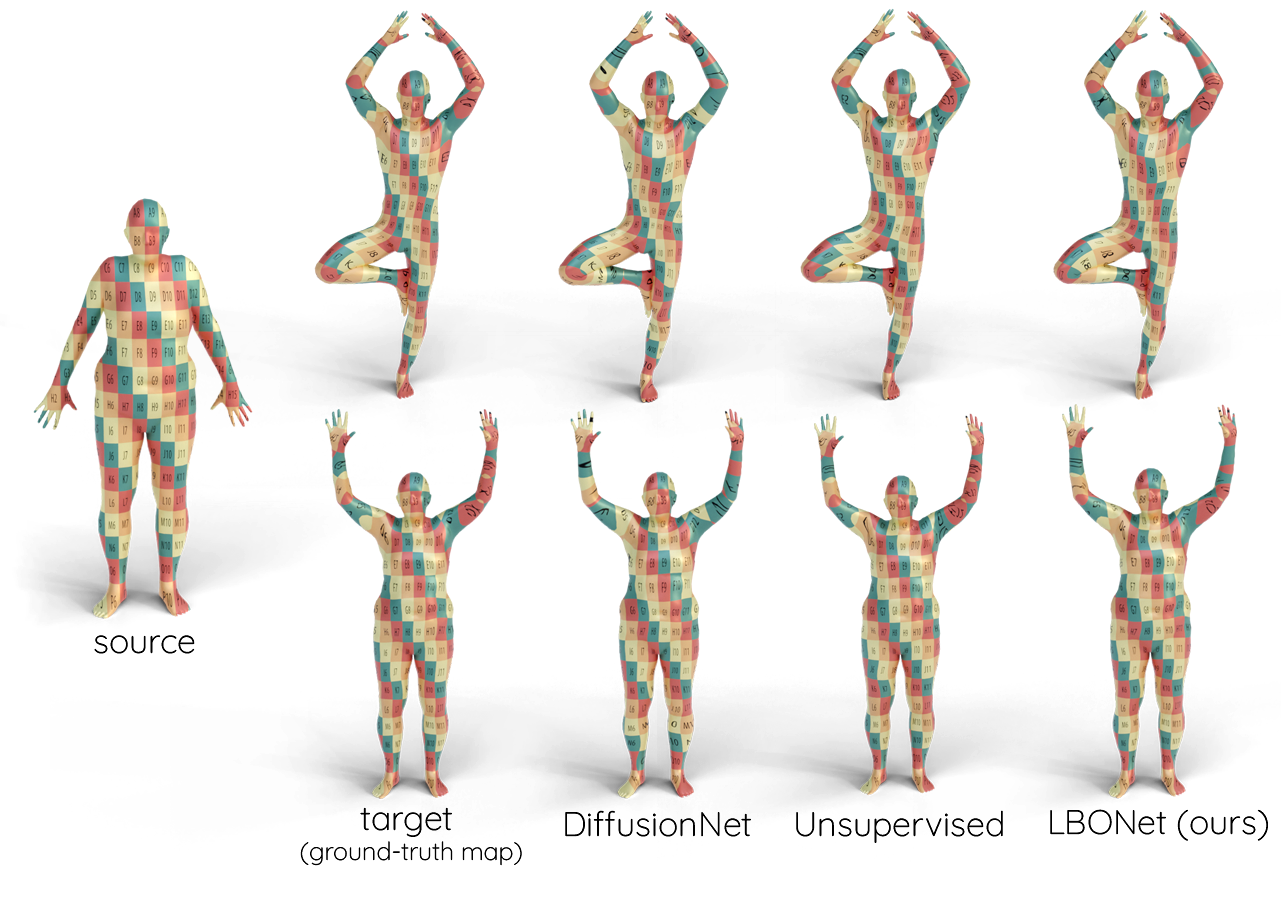}

    \caption{Texture transfer experiment on the FAUST test set. Source illustrates the texture applied to the source shape, and target the texture mapped via the ground-truth correspondence to the target shape. Followed by the illustrations of the predictions of DiffusionNet, Unsupervised, and LBONet.}
            \label{fig:coarse}
\end{figure*}
In the texture transfer illustration shown in Figure \ref{fig:coarse}, LBONet achieves the best performance in regions where significant stretching occurs (top row: legs; bottom row: armpits), outperforming both DiffusionNet and the unsupervised baseline, which struggle in these areas. Two minor artifacts are observable in LBONet's output: the texture patches are slightly distorted compared to the ground truth, and there is some misalignment in the left forearm. This tradeoff is characteristic of LBONet’s behavior across various tasks. It demonstrates that LBONet effectively transforms the static LBO into a learnable optimization space, where improved solutions become attainable. However, this flexibility also introduces challenges: While better solutions may exist, the resulting loss landscape can hinder convergence to globally optimal results.

Nevertheless, LBONet outperforms Unsupervised and correctly recognizes stretched areas.

\subsection{Ablation Study}
To prove the effectiveness of LBONet, we repeat the human segmentation \cite{maron2017convolutional} experiment in a more traditional setting. We conduct the experiment by using the spectral descriptor directly (see Figure \ref{fig:arch} Frontend) with a single shared MLP instead of a Backend. This way it is down to the bare minimum and we can measure LBONets contribution almost in isolation. Table \ref{tab:ablationtable} shows the results of each layer. Several observations can be made. VoronoiNet on its own does not provide much benefit, however when run in combination of the other networks such as RiemannNet or ALBONet it creates a synergistic effect. RiemannNet and ALBONet seems to have some overlapping effect as they still add some performance, however the performance gain diminishes. The most intriguing effect however is, that when ALBO is allowed to learn the anisotropic factor in both directions (ALBO+), we see a substantial increase in performance, which in the literature and methods using ALBO went mostly unnoticed. 

\begin{table}[!th]

\begin{center}
\setlength{\tabcolsep}{2pt}
\begin{tabular}{ cccccc}
 RiemannNet & ALBONet & ALBO+Net  & VoronoiNet & Accuracy & Improvement\\
\midrule

- & - & - & - & 84.1 &    \\ 
- & - & - & \checkmark & 84.6 & 0.6\%  \\
- & \checkmark & - & - & 85.6 & 1.78\%\\
\checkmark & - & - & - & 86.9& 3.33\%  \\
- & \checkmark & - & \checkmark & 87.6 & 4.16\% \\
\checkmark & \checkmark & - & - & 87.8 & 4.40\% \\
- & - & \checkmark & - & 88.1 & 4.76\%\\
\checkmark & - & - & \checkmark & 88.1 & 4.76\% \\
\checkmark & \checkmark & - & \checkmark & 88.3 & 4.99\% \\
\checkmark & - & \checkmark & - & 89.1 & 5.95\% \\
- & - & \checkmark & \checkmark & 89.9 & 6.90\%\\
\checkmark & - & \checkmark & \checkmark & \textbf{90.7} & 7.85\% \\
\midrule
\end{tabular}

\caption{\label{tab:ablationtable}Results on the human segmentation benchmark. The table reports the performance of the proposed modules, both individually and in combination. A \checkmark\ symbol indicates that the corresponding module was used to obtain the reported result.}
\end{center}
\end{table}

\begin{figure}
    \centering

    \includegraphics[scale=0.3,trim=3cm 2cm 0cm 0cm]{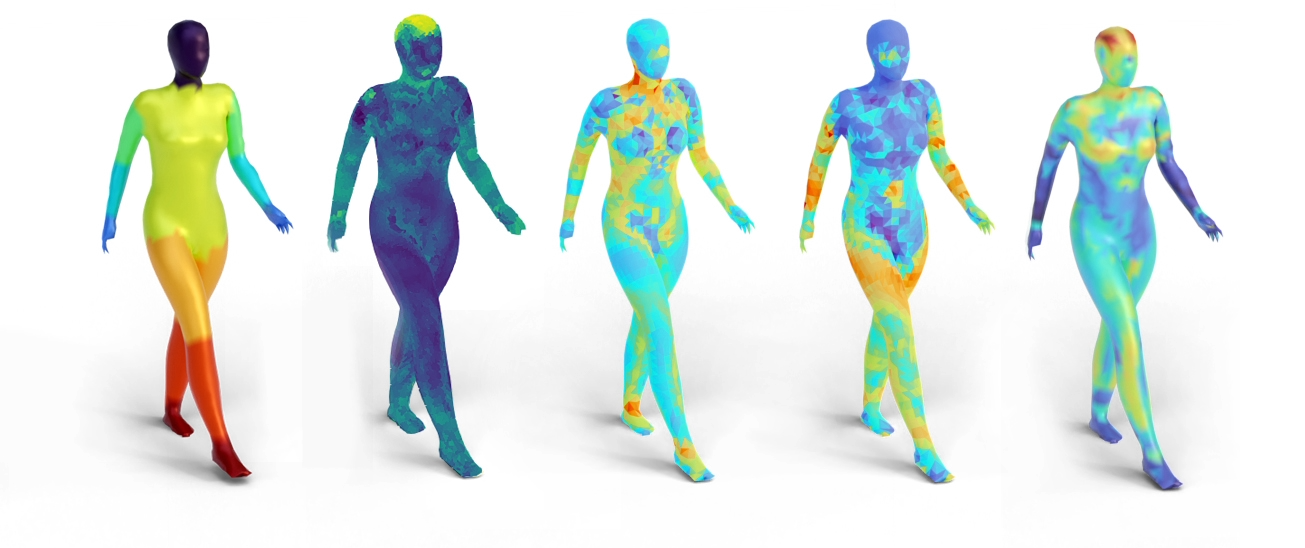}
\caption{Ground-truth segmentation of a manifold (first column) alongside learned features on a test manifold from the human segmentation benchmark. Results are shown for RiemannNet (second), ALBONet (third and fourth), and VoronoiNet (fifth). Distinctive patterns emerge along the segmentation boundaries, highlighting that different modules exhibit strengths in different aspects of the task.}
            \label{fig:humanssegfeatures}
\end{figure}

\begin{figure}
    \centering

    \includegraphics[scale=0.1,trim=0cm 3cm 5cm 0cm]{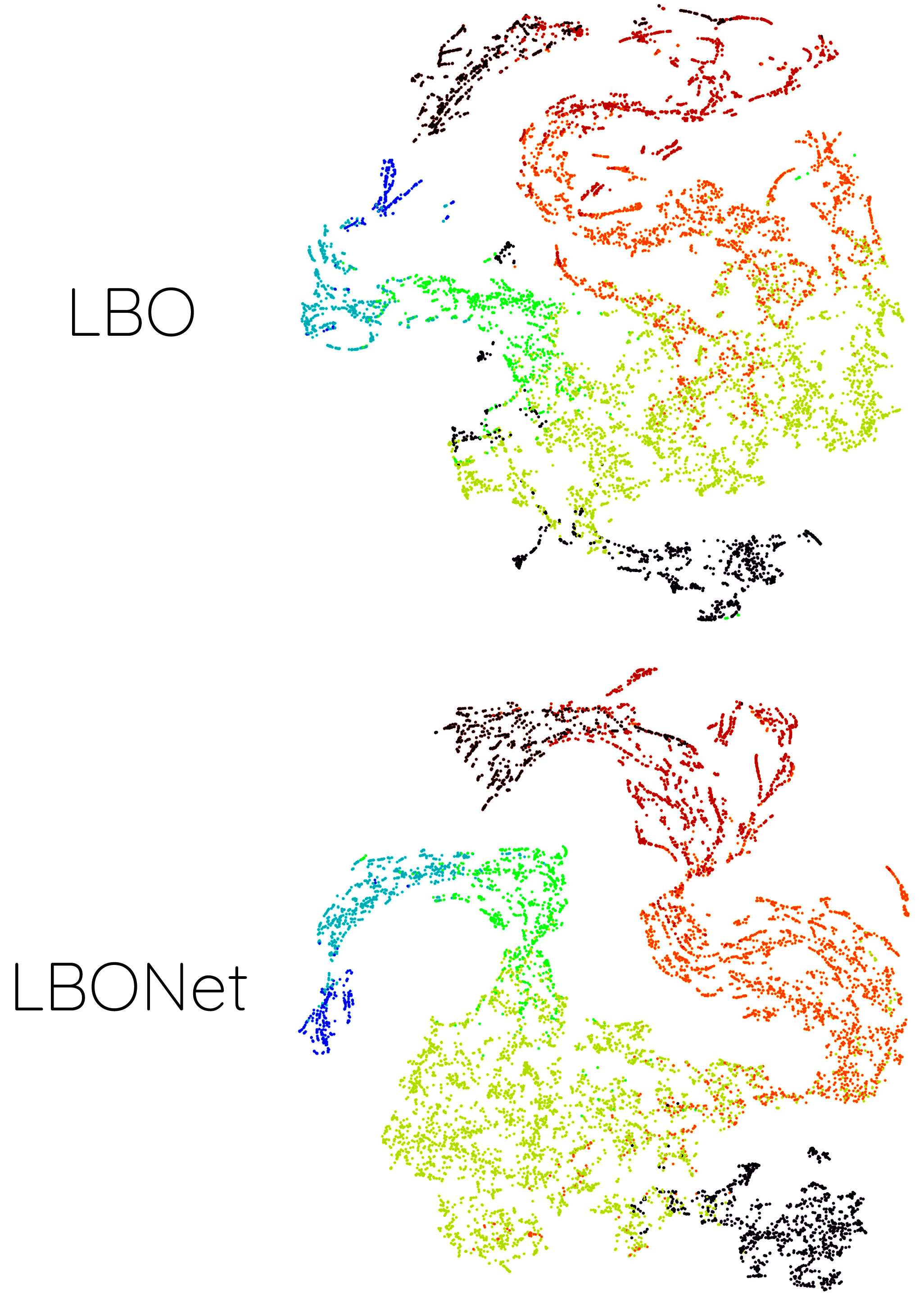}
\caption{Heat kernel signature (HKS) features obtained from the ablation study presented in Chapter 5. The top row shows HKS features computed using the standard LBO, while the bottom row shows those computed using LBONet. Features were reduced from 16 dimensions to 2D using t-SNE for visualization.}
            \label{fig:tsne}
\end{figure}

\label{sec:ablation}
To further understand how LBONet is learning in a specific task, we visualize the learned features in a segmentation task in Figure \ref{fig:humanssegfeatures}. At this point it is important to differentiate that LBONet can be used with any spectral descriptor, but when used with the heat kernel signature, the spectrum will be optimized wrt. to what the signature picks out from the eigenbasis. Hence, in a correspondence task we can expect the eigenbasis to be closer after learning with LBONet, however in a segmentation task that is not necessarily the case, as LBONet will learn an eigenbasis which allows better separation of segments. As interpretation of learned features within a neural network is a difficult task and might not always yield to explainable results, some patterns are evident in Figure \ref{fig:humanssegfeatures}. This becomes clear when considering the ground-truth segmentation. Firstly, when observing the legs, we see that ALBONet reduces the anisotropic factor into the maximum curvature direction and applies some rotation, which acts as a boundary in the spectral space between the torso and the leg. Secondly, we see a high activity on the head with RiemannNet and VoronoiNet, which both enlarge the head in the spectral space. By enlarging the head in the spectral space, the head is becoming easier to distinguish from the torso and upper arm. Lastly, we see scattered detections of further boundaries, which help to to separate the segments even more. The t-SNE \cite{van2008visualizing} method allows to visualize high-dimensional data in lower dimensions. By this, we can compare the standard heat kernel signature and the LBONet heat kernel signature. We visualize randomly sampled points of the signature on the test set colored by their ground-truth labels, which can be observed in Figure \ref{fig:tsne}. Several observations can be made from the distribution of the features and aligns with the hypothesis made earlier. We can observe that the standard LBO misclassifies lots of head parts either as torso or even as upper arms, which LBONet is able to eliminate completely. The overlap between the torso and the leg is reduced by LBONet significantly, only allowing a small part to overlap. Furthermore, we see that each segment boundary has been narrowed down drastically, which allows easy separation of segments in the spectral space, which with the standard LBO seems like a hard task.

\begin{figure}[ht]
    \centering
    \begin{tabular}{c|c}
    \hspace{-2cm}
     \newcommand{\pckLineWidth}{2pt}
\newcommand{\plotWidth}{\columnwidth}
\newcommand{\plotHeight}{0.7\columnwidth}
\newcommand{\pckTitle}{\textbf{SHREC’14 Real }}
\definecolor{cPLOT0}{RGB}{28,213,227}
\definecolor{cPLOT1}{RGB}{80,150,80}
\definecolor{cPLOT2}{RGB}{90,130,213}
\definecolor{cPLOT3}{RGB}{247,179,43}
\definecolor{cPLOT5}{RGB}{242,64,0}

\pgfplotsset{%
    label style = {font=\large},
    tick label style = {font=\large},
    title style =  {font=\LARGE},
    legend style={  fill= gray!10,
                    fill opacity=0.6, 
                    font=\large,
                    draw=gray!20, %
                    text opacity=1}
}
\begin{tikzpicture}[scale=0.55, transform shape]
	\begin{axis}[
		width=\plotWidth,
		height=\plotHeight,
		grid=major,
		title=\pckTitle,
		legend style={
			at={(0.53,0.03)},
			anchor=south east,
			legend columns=1},
		legend cell align={left},
        xlabel={Recall},
        ylabel={Precision},
		xmin=0.1,
        xmax=1.0,
        ylabel near ticks,
        xtick={0.1, 0.2, 0.3, 0.4, 0.5, 0.6, 0.7, 0.8, 0.9, 1.0},
	ymin=0.4,
        ymax=1,
        ytick={0, 0.20, 0.40, 0.60, 0.80, 1.0}
	]

\addplot [color=cPLOT1, smooth, line width=\pckLineWidth]
table[row sep=crcr]{%
0.1000  1.0000\\
0.1500  0.9662\\
0.2000  0.9324\\
0.2500  0.9207\\
0.3000  0.9089\\
0.3500  0.8988\\
0.4000  0.8887\\
0.4500  0.8794\\
0.5000  0.8702\\
0.5500  0.8643\\
0.6000  0.8585\\
0.6500  0.8503\\
0.7000  0.8421\\
0.7500  0.8247\\
0.8000  0.8072\\
0.8500  0.7885\\
0.9000  0.7697\\
0.9500  0.7343\\
1.0000  0.6989\\
    };
\addlegendentry{\textcolor{black}{LBONet}}

\addplot [color=cPLOT2, smooth, line width=\pckLineWidth]
table[row sep=crcr]{%
0.1000 1.0000 \\
0.1500 0.9530 \\
0.2000 0.9060 \\
0.2500 0.8885 \\
0.3000 0.8711 \\
0.3500 0.8485 \\
0.4000 0.8259 \\
0.4500 0.8138 \\
0.5000 0.8018 \\
0.5500 0.7913 \\
0.6000 0.7808 \\
0.6500 0.7624 \\
0.7000 0.7441 \\
0.7500 0.7239 \\
0.8000 0.7037 \\
0.8500 0.6771 \\
0.9000 0.6504 \\
0.9500 0.6054 \\
1.0000 0.5603 \\
    };
\addlegendentry{\textcolor{black}{Baseline}}  

\addplot [color=cPLOT5, smooth, line width=\pckLineWidth]
table[row sep=crcr]{%
0.1000 1.0000 \\
0.1500 0.9108 \\
0.2000 0.8217 \\
0.2500 0.7924 \\
0.3000 0.7632 \\
0.3500 0.7401 \\
0.4000 0.7170 \\
0.4500 0.7032 \\
0.5000 0.6894 \\
0.5500 0.6766 \\
0.6000 0.6639 \\
0.6500 0.6535 \\
0.7000 0.6432 \\
0.7500 0.6259 \\
0.8000 0.6086 \\
0.8500 0.5870 \\
0.9000 0.5654 \\
0.9500 0.5317 \\
1.0000 0.4980 \\
    };
\addlegendentry{\textcolor{black}{LBONet (HKS)}} 

\addplot [color=cPLOT0, smooth, line width=\pckLineWidth]
table[row sep=crcr]{%
0.1000 1.0000 \\
0.1500 0.8928 \\
0.2000 0.7856 \\
0.2500 0.7499 \\
0.3000 0.7142 \\
0.3500 0.6865 \\
0.4000 0.6587 \\
0.4500 0.6476 \\
0.5000 0.6365 \\
0.5500 0.6229 \\
0.6000 0.6092 \\
0.6500 0.5963 \\
0.7000 0.5834 \\
0.7500 0.5601 \\
0.8000 0.5368 \\
0.8500 0.5135 \\
0.9000 0.4901 \\
0.9500 0.4455 \\
1.0000 0.4009 \\
    };
\addlegendentry{\textcolor{black}{Baseline (HKS)}}

\end{axis}
\end{tikzpicture}&
     \hspace{-1.6cm}
     \newcommand{\pckLineWidth}{2pt}
\newcommand{\plotWidth}{\columnwidth}
\newcommand{\plotHeight}{0.7\columnwidth}
\newcommand{\pckTitle}{\textbf{SHREC’14 Synthetic }}
\definecolor{cPLOT0}{RGB}{28,213,227}
\definecolor{cPLOT1}{RGB}{80,150,80}
\definecolor{cPLOT2}{RGB}{90,130,213}
\definecolor{cPLOT3}{RGB}{247,179,43}
\definecolor{cPLOT5}{RGB}{242,64,0}

\pgfplotsset{%
    label style = {font=\large},
    tick label style = {font=\large},
    title style =  {font=\LARGE},
    legend style={  fill= gray!10,
                    fill opacity=0.6, 
                    font=\large,
                    draw=gray!20, %
                    text opacity=1}
}
\begin{tikzpicture}[scale=0.55, transform shape]
	\begin{axis}[
		width=\plotWidth,
		height=\plotHeight,
		grid=major,
		title=\pckTitle,
		legend style={
			at={(0.53,0.03)},
			anchor=south east,
			legend columns=1},
		legend cell align={left},
        xlabel={Recall},
		xmin=0.1,
        xmax=1.0,
        ylabel near ticks,
        xtick={0.1, 0.2, 0.3, 0.4, 0.5, 0.6, 0.7, 0.8, 0.9, 1.0},
	ymin=0.5,
        ymax=1,
        ytick={0, 0.20, 0.40, 0.60, 0.80, 1.0}
	]

\addplot [color=cPLOT1, smooth, line width=\pckLineWidth]
table[row sep=crcr]{%
0.050 1 \\
0.100 0.995952 \\
0.150 0.9874 \\
0.200 0.9858 \\
0.250 0.9824 \\
0.300 0.9779 \\
0.350 0.9753 \\
0.400 0.9712 \\
0.450 0.9681 \\
0.500 0.9623 \\
0.550 0.9591 \\
0.600 0.9586 \\
0.650 0.9520 \\
0.700 0.9450 \\
0.750 0.9379 \\
0.800 0.9252 \\
0.850 0.9131 \\
0.900 0.8802 \\
0.950 0.8257 \\
1.000 0.7548 \\
    };
\addlegendentry{\textcolor{black}{LBONet}}

\addplot [color=cPLOT2, smooth, line width=\pckLineWidth]
table[row sep=crcr]{%
0.05 1 \\
0.100 0.98711 \\
0.150 0.9693 \\
0.200 0.9591 \\
0.250 0.9448 \\
0.300 0.9412 \\
0.350 0.9317 \\
0.400 0.9269 \\
0.450 0.9211 \\
0.500 0.9150 \\
0.550 0.9050 \\
0.600 0.8991 \\
0.650 0.8892 \\
0.700 0.8777 \\
0.750 0.8667 \\
0.800 0.8547 \\
0.850 0.8375 \\
0.900 0.7774 \\
0.950 0.6796 \\
1.000 0.6151 \\
    };
\addlegendentry{\textcolor{black}{Baseline}}  

\addplot [color=cPLOT5, smooth, line width=\pckLineWidth]
table[row sep=crcr]{%
0.05 1 \\
0.100 0.913316 \\
0.150 0.8886 \\
0.200 0.8824 \\
0.250 0.8680 \\
0.300 0.8578 \\
0.350 0.8454 \\
0.400 0.8351 \\
0.450 0.8290 \\
0.500 0.8250 \\
0.550 0.8157 \\
0.600 0.8069 \\
0.650 0.8000 \\
0.700 0.7890 \\
0.750 0.7756 \\
0.800 0.7580 \\
0.850 0.7352 \\
0.900 0.7096 \\
0.950 0.6711 \\
1.000 0.6261 \\
    };
\addlegendentry{\textcolor{black}{LBONet (HKS)}} 

\addplot [color=cPLOT0, smooth, line width=\pckLineWidth]
table[row sep=crcr]{%
0.05 1 \\
0.100 0.885731 \\
0.150 0.8481 \\
0.200 0.8326 \\
0.250 0.8137 \\
0.300 0.7963 \\
0.350 0.7755 \\
0.400 0.7614 \\
0.450 0.7483 \\
0.500 0.7392 \\
0.550 0.7298 \\
0.600 0.7172 \\
0.650 0.7086 \\
0.700 0.6979 \\
0.750 0.6844 \\
0.800 0.6697 \\
0.850 0.6551 \\
0.900 0.6284 \\
0.950 0.5795 \\
1.000 0.5131 \\
    };
\addlegendentry{\textcolor{black}{Baseline (HKS)}}

\end{axis}
\end{tikzpicture}
    \end{tabular}
    \caption{{ Precision–Recall curves for LBONet, the baseline, LBONet (HKS), and LBO (HKS), evaluated on the SHREC'14 Real benchmark (top) and the SHREC'14 Synthetic benchmark (bottom). Distances were computed using Euclidean distance. The performance of the spectral signatures for both LBONet (HKS) and LBO (HKS) was evaluated using average pooling.}}
    \label{fig:shrecret}
\end{figure}
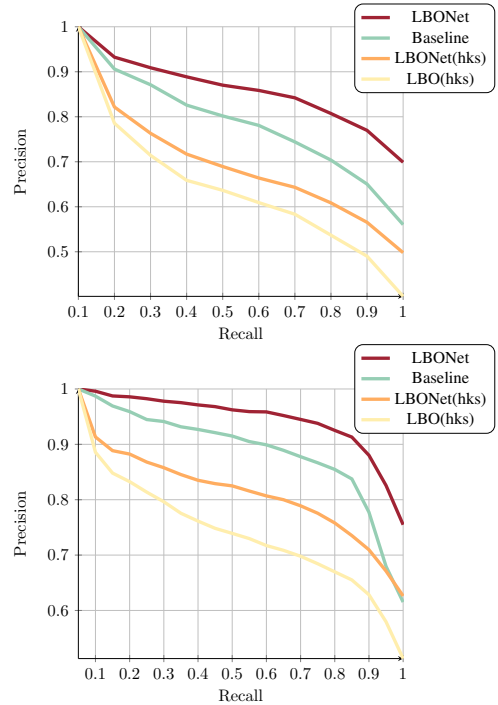

To understand the gain in a retrieval task we compare the standard heat kernel signature to the LBONet heat kernel signature by average pooling. The performance is plotted in Figure \ref{fig:shrecret} alongside their performance achieved in the experiment \ref{tab:res14} and \ref{tab:res14s}. One can observe that the performance gain in the final experiment was due to LBONet as the heat kernel signature with LBONet performs better than the standard heat kernel signature and that this performance gain translates into the final result. Similar to the ablation done in the segmentation task, we can observe that the LBONet heat kernel signature is far more descriptive.


\section{Conclusion}
We have introduced a framework, which allows to adapt the LBO eigenbasis to the task at hand without precalculating LBO. The modules RiemanNet, ALBONet and VoronoiNet are capable of learning powerful features and add provably value to the descriptiveness of spectral descriptors. The experiments demonstrated that LBONet is versatile and can make the LBO robust against near-isometries and help to separate segments in spectral space. Moreover, we have addressed the problem of scalability, which remains still a relevant topic for future research.

\section{Future work}
Future research could extend LBONet by incorporating additional modules or exploring related techniques such as isospectralization. One promising direction is the integration of the recently introduced FLBO \cite{weber2024finsler}, which could enable LBONet to learn and represent asymmetries on the manifold. Another important avenue is improving scalability. Although LBONet is effective, backpropagation currently requires operations involving large, dense matrices, which limits its applicability to high-resolution meshes. Additionally, handling imperfect or incomplete geometry remains a significant challenge. In particular, the method cannot yet be applied to partial shapes or meshes with holes, as the eigenbasis becomes highly unstable under such conditions, making it difficult to learn generalizable features across shapes. Addressing this instability is essential for extending LBONet to broader real-world scenarios involving partial, noisy, or corrupted geometric data.

\newpage

\bibliographystyle{IEEEtran}
\bibliography{main}

\begin{thebibliography}{10}
\providecommand{\url}[1]{#1}
\csname url@samestyle\endcsname
\providecommand{\newblock}{\relax}
\providecommand{\bibinfo}[2]{#2}
\providecommand{\BIBentrySTDinterwordspacing}{\spaceskip=0pt\relax}
\providecommand{\BIBentryALTinterwordstretchfactor}{4}
\providecommand{\BIBentryALTinterwordspacing}{\spaceskip=\fontdimen2\font plus
\BIBentryALTinterwordstretchfactor\fontdimen3\font minus \fontdimen4\font\relax}
\providecommand{\BIBforeignlanguage}[2]{{%
\expandafter\ifx\csname l@#1\endcsname\relax
\typeout{** WARNING: IEEEtran.bst: No hyphenation pattern has been}%
\typeout{** loaded for the language `#1'. Using the pattern for}%
\typeout{** the default language instead.}%
\else
\language=\csname l@#1\endcsname
\fi
#2}}
\providecommand{\BIBdecl}{\relax}
\BIBdecl

\bibitem{kokkinos2012intrinsic}
I.~Kokkinos, M.~M. Bronstein, R.~Litman, and A.~M. Bronstein, ``Intrinsic shape context descriptors for deformable shapes,'' in \emph{2012 IEEE Conference on Computer Vision and Pattern Recognition}.\hskip 1em plus 0.5em minus 0.4em\relax IEEE, 2012, pp. 159--166.

\bibitem{litman2014supervised}
R.~Litman, A.~Bronstein, M.~Bronstein, and U.~Castellani, ``Supervised learning of bag-of-features shape descriptors using sparse coding,'' in \emph{Computer Graphics Forum}, vol.~33, no.~5.\hskip 1em plus 0.5em minus 0.4em\relax Wiley Online Library, 2014, pp. 127--136.

\bibitem{masci2015shapenet}
J.~Masci, D.~Boscaini, M.~Bronstein, and P.~Vandergheynst, ``Shapenet: Convolutional neural networks on non-euclidean manifolds,'' Tech. Rep., 2015.

\bibitem{yi2017syncspeccnn}
L.~Yi, H.~Su, X.~Guo, and L.~J. Guibas, ``Syncspeccnn: Synchronized spectral cnn for 3d shape segmentation,'' in \emph{Proceedings of the IEEE Conference on Computer Vision and Pattern Recognition}, 2017, pp. 2282--2290.

\bibitem{qi2017pointnet++}
C.~R. Qi, L.~Yi, H.~Su, and L.~J. Guibas, ``Pointnet++: Deep hierarchical feature learning on point sets in a metric space,'' in \emph{Advances in neural information processing systems}, 2017, pp. 5099--5108.

\bibitem{yu2020second}
R.~Yu, J.~Sun, and H.~Li, ``Second-order spectral transform block for 3d shape classification and retrieval,'' \emph{IEEE Transactions on Image Processing}, vol.~29, pp. 4530--4543, 2020.

\bibitem{sharp2022diffusionnet}
N.~Sharp, S.~Attaiki, K.~Crane, and M.~Ovsjanikov, ``Diffusionnet: Discretization agnostic learning on surfaces,'' \emph{ACM Transactions on Graphics (TOG)}, vol.~41, no.~3, pp. 1--16, 2022.

\bibitem{limberger2015feature}
F.~A. Limberger and R.~C. Wilson, ``Feature encoding of spectral signatures for 3d non-rigid shape retrieval.'' 2015.

\bibitem{giachetti2012radial}
A.~Giachetti and C.~Lovato, ``Radial symmetry detection and shape characterization with the multiscale area projection transform,'' in \emph{Computer graphics forum}, vol.~31, no.~5.\hskip 1em plus 0.5em minus 0.4em\relax Wiley Online Library, 2012, pp. 1669--1678.

\bibitem{ye2013fast}
J.~Ye, Z.~Yan, and Y.~Yu, ``Fast nonrigid 3d retrieval using modal space transform,'' in \emph{Proceedings of the 3rd ACM conference on International conference on multimedia retrieval}, 2013, pp. 121--126.

\bibitem{limberger2018curvature}
F.~A. Limberger and R.~C. Wilson, ``Curvature-based spectral signatures for non-rigid shape retrieval,'' \emph{Computer Vision and Image Understanding}, vol. 172, pp. 1--11, 2018.

\bibitem{choukroun2018hamiltonian}
Y.~Choukroun, A.~Shtern, A.~Bronstein, and R.~Kimmel, ``Hamiltonian operator for spectral shape analysis,'' \emph{IEEE transactions on visualization and computer graphics}, vol.~26, no.~2, pp. 1320--1331, 2018.

\bibitem{andreux2015anisotropic}
M.~Andreux, E.~Rodola, M.~Aubry, and D.~Cremers, ``Anisotropic laplace-beltrami operators for shape analysis,'' in \emph{Computer Vision-ECCV 2014 Workshops: Zurich, Switzerland, September 6-7 and 12, 2014, Proceedings, Part IV 13}.\hskip 1em plus 0.5em minus 0.4em\relax Springer, 2015, pp. 299--312.

\bibitem{smirnov2021hodgenet}
D.~Smirnov and J.~Solomon, ``Hodgenet: Learning spectral geometry on triangle meshes,'' \emph{ACM Transactions on Graphics (TOG)}, vol.~40, no.~4, pp. 1--11, 2021.

\bibitem{reuter2005laplace}
M.~Reuter, F.-E. Wolter, and N.~Peinecke, ``Laplace-spectra as fingerprints for shape matching,'' in \emph{Proceedings of the 2005 ACM symposium on Solid and physical modeling}.\hskip 1em plus 0.5em minus 0.4em\relax ACM, 2005, pp. 101--106.

\bibitem{meyer2003discrete}
M.~Meyer, M.~Desbrun, P.~Schr{\"o}der, and A.~H. Barr, ``Discrete differential-geometry operators for triangulated 2-manifolds,'' in \emph{Visualization and mathematics III}.\hskip 1em plus 0.5em minus 0.4em\relax Springer, 2003, pp. 35--57.

\bibitem{gahm2018riemannian}
J.~K. Gahm, Y.~Shi, A.~D.~N. Initiative \emph{et~al.}, ``Riemannian metric optimization on surfaces (rmos) for intrinsic brain mapping in the laplace--beltrami embedding space,'' \emph{Medical image analysis}, vol.~46, pp. 189--201, 2018.

\bibitem{rustamov2007laplace}
R.~M. Rustamov, ``Laplace-beltrami eigenfunctions for deformation invariant shape representation,'' in \emph{Proceedings of the fifth Eurographics symposium on Geometry processing}.\hskip 1em plus 0.5em minus 0.4em\relax Eurographics Association, 2007, pp. 225--233.

\bibitem{sun2009concise}
J.~Sun, M.~Ovsjanikov, and L.~Guibas, ``A concise and provably informative multi-scale signature based on heat diffusion,'' in \emph{Computer graphics forum}, vol.~28, no.~5.\hskip 1em plus 0.5em minus 0.4em\relax Wiley Online Library, 2009, pp. 1383--1392.

\bibitem{bronstein2010scale}
M.~M. Bronstein and I.~Kokkinos, ``Scale-invariant heat kernel signatures for non-rigid shape recognition,'' in \emph{2010 IEEE Computer Society Conference on Computer Vision and Pattern Recognition}.\hskip 1em plus 0.5em minus 0.4em\relax IEEE, 2010, pp. 1704--1711.

\bibitem{naffouti2018heuristic}
S.~E. Naffouti, Y.~Fougerolle, I.~Aouissaoui, A.~Sakly, and F.~M{\'e}riaudeau, ``Heuristic optimization-based wave kernel descriptor for deformable 3d shape matching and retrieval,'' \emph{Signal, Image and Video Processing}, vol.~12, no.~5, pp. 915--923, 2018.

\bibitem{litman2013learning}
R.~Litman and A.~M. Bronstein, ``Learning spectral descriptors for deformable shape correspondence,'' \emph{IEEE transactions on pattern analysis and machine intelligence}, vol.~36, no.~1, pp. 171--180, 2013.

\bibitem{wang2019dynamic}
Y.~Wang, Y.~Sun, Z.~Liu, S.~E. Sarma, M.~M. Bronstein, and J.~M. Solomon, ``Dynamic graph cnn for learning on point clouds,'' \emph{Acm Transactions On Graphics (tog)}, vol.~38, no.~5, pp. 1--12, 2019.

\bibitem{boscaini2015learning}
D.~Boscaini, J.~Masci, S.~Melzi, M.~M. Bronstein, U.~Castellani, and P.~Vandergheynst, ``Learning class-specific descriptors for deformable shapes using localized spectral convolutional networks,'' in \emph{Computer Graphics Forum}, vol.~34, no.~5.\hskip 1em plus 0.5em minus 0.4em\relax Wiley Online Library, 2015, pp. 13--23.

\bibitem{boscaini2016anisotropic}
D.~Boscaini, J.~Masci, E.~Rodol{\`a}, M.~M. Bronstein, and D.~Cremers, ``Anisotropic diffusion descriptors,'' in \emph{Computer Graphics Forum}, vol.~35, no.~2.\hskip 1em plus 0.5em minus 0.4em\relax Wiley Online Library, 2016, pp. 431--441.

\bibitem{weber2024finsler}
S.~Weber, T.~Dag{\`e}s, M.~Gao, and D.~Cremers, ``Finsler-laplace-beltrami operators with application to shape analysis,'' \emph{arXiv preprint arXiv:2404.03999}, 2024.

\bibitem{xie2016learned}
J.~Xie, M.~Wang, and Y.~Fang, ``Learned binary spectral shape descriptor for 3d shape correspondence,'' in \emph{Proceedings of the IEEE conference on computer vision and pattern recognition}, 2016, pp. 3309--3317.

\bibitem{xie2015deepshape}
J.~Xie, Y.~Fang, F.~Zhu, and E.~Wong, ``Deepshape: Deep learned shape descriptor for 3d shape matching and retrieval,'' in \emph{Proceedings of the IEEE Conference on Computer Vision and Pattern Recognition}, 2015, pp. 1275--1283.

\bibitem{cosmo2019isospectralization}
L.~Cosmo, M.~Panine, A.~Rampini, M.~Ovsjanikov, M.~M. Bronstein, and E.~Rodola, ``Isospectralization, or how to hear shape, style, and correspondence,'' in \emph{Proceedings of the IEEE/CVF conference on computer vision and pattern recognition}, 2019, pp. 7529--7538.

\bibitem{rampini2019correspondence}
A.~Rampini, I.~Tallini, M.~Ovsjanikov, A.~M. Bronstein, and E.~Rodola, ``Correspondence-free region localization for partial shape similarity via hamiltonian spectrum alignment,'' in \emph{2019 International Conference on 3D Vision (3DV)}.\hskip 1em plus 0.5em minus 0.4em\relax IEEE, 2019, pp. 37--46.

\bibitem{shi2011conformal}
Y.~Shi, R.~Lai, R.~Gill, D.~Pelletier, D.~Mohr, N.~Sicotte, and A.~W. Toga, ``Conformal metric optimization on surface (cmos) for deformation and mapping in laplace-beltrami embedding space,'' in \emph{International Conference on Medical Image Computing and Computer-Assisted Intervention}.\hskip 1em plus 0.5em minus 0.4em\relax Springer, 2011, pp. 327--334.

\bibitem{ovsjanikov2012functional}
M.~Ovsjanikov, M.~Ben-Chen, J.~Solomon, A.~Butscher, and L.~Guibas, ``Functional maps: a flexible representation of maps between shapes,'' \emph{ACM Transactions on Graphics (ToG)}, vol.~31, no.~4, pp. 1--11, 2012.

\bibitem{aubry2011wave}
M.~Aubry, U.~Schlickewei, and D.~Cremers, ``The wave kernel signature: A quantum mechanical approach to shape analysis,'' in \emph{2011 IEEE international conference on computer vision workshops (ICCV workshops)}.\hskip 1em plus 0.5em minus 0.4em\relax IEEE, 2011, pp. 1626--1633.

\bibitem{cosmo20223d}
L.~Cosmo, G.~Minello, M.~Bronstein, E.~Rodol{\`a}, L.~Rossi, and A.~Torsello, ``3d shape analysis through a quantum lens: the average mixing kernel signature,'' \emph{International Journal of Computer Vision}, vol. 130, no.~6, pp. 1474--1493, 2022.

\bibitem{qi2017pointnet}
C.~R. Qi, H.~Su, K.~Mo, and L.~J. Guibas, ``Pointnet: Deep learning on point sets for 3d classification and segmentation,'' in \emph{Proceedings of the IEEE conference on computer vision and pattern recognition}, 2017, pp. 652--660.

\bibitem{guo2021pct}
M.-H. Guo, J.-X. Cai, Z.-N. Liu, T.-J. Mu, R.~R. Martin, and S.-M. Hu, ``Pct: Point cloud transformer,'' \emph{Computational Visual Media}, vol.~7, pp. 187--199, 2021.

\bibitem{wu2022point}
H.~Zhao, L.~Jiang, J.~Jia, P.~H. Torr, and V.~Koltun, ``Point transformer,'' in \emph{Proceedings of the IEEE/CVF international conference on computer vision}, 2021, pp. 16\,259--16\,268.

\bibitem{wong2023heat}
C.-C. Wong, ``Heat diffusion based multi-scale and geometric structure-aware transformer for mesh segmentation,'' in \emph{Proceedings of the IEEE/CVF Conference on Computer Vision and Pattern Recognition}, 2023, pp. 4413--4422.

\bibitem{rusinkiewicz2004estimating}
S.~Rusinkiewicz, ``Estimating curvatures and their derivatives on triangle meshes,'' in \emph{Proceedings. 2nd International Symposium on 3D Data Processing, Visualization and Transmission, 2004. 3DPVT 2004.}\hskip 1em plus 0.5em minus 0.4em\relax IEEE, 2004, pp. 486--493.

\bibitem{tang2012evaluation}
S.~Tang and A.~Godil, ``An evaluation of local shape descriptors for 3d shape retrieval,'' in \emph{Three-Dimensional Image Processing (3DIP) and Applications II}, vol. 8290.\hskip 1em plus 0.5em minus 0.4em\relax International Society for Optics and Photonics, 2012, p. 82900N.

\bibitem{nelson1976simplified}
R.~B. Nelson, ``Simplified calculation of eigenvector derivatives,'' \emph{AIAA journal}, vol.~14, no.~9, pp. 1201--1205, 1976.

\bibitem{rosen1960gradient}
J.~B. Rosen, ``The gradient projection method for nonlinear programming. part i. linear constraints,'' \emph{Journal of the society for industrial and applied mathematics}, vol.~8, no.~1, pp. 181--217, 1960.

\bibitem{DhillonST04}
\BIBentryALTinterwordspacing
I.~S. Dhillon, S.~Sra, and J.~A. Tropp, ``Triangle fixing algorithms for the metric nearness problem,'' in \emph{NIPS}, 2004, pp. 361--368. [Online]. Available: \url{http://papers.nips.cc/paper/2598-triangle-fixing-algorithms-for-the-metric-nearness-problem}
\BIBentrySTDinterwordspacing

\bibitem{boscaini2016learning}
D.~Boscaini, J.~Masci, E.~Rodol{\`a}, and M.~Bronstein, ``Learning shape correspondence with anisotropic convolutional neural networks,'' \emph{Advances in neural information processing systems}, vol.~29, 2016.

\bibitem{li2020shape}
Q.~Li, S.~Liu, L.~Hu, and X.~Liu, ``Shape correspondence using anisotropic chebyshev spectral cnns,'' in \emph{Proceedings of the IEEE/CVF conference on Computer Vision and Pattern Recognition}, 2020, pp. 14\,658--14\,667.

\bibitem{hanocka2019meshcnn}
R.~Hanocka, A.~Hertz, N.~Fish, R.~Giryes, S.~Fleishman, and D.~Cohen-Or, ``Meshcnn: a network with an edge,'' \emph{ACM Transactions on Graphics (TOG)}, vol.~38, no.~4, pp. 1--12, 2019.

\bibitem{Virtanen2020}
\BIBentryALTinterwordspacing
P.~Virtanen, R.~Gommers, T.~E. Oliphant, M.~Haberland, T.~Reddy, D.~Cournapeau, E.~Burovski, P.~Peterson, W.~Weckesser, J.~Bright, S.~J. van~der Walt, M.~Brett, J.~Wilson, K.~J. Millman, N.~Mayorov, A.~R.~J. Nelson, E.~Jones, R.~Kern, E.~Larson, C.~J. Carey, {\.{I}}.~Polat, Y.~Feng, E.~W. Moore, J.~VanderPlas, D.~Laxalde, J.~Perktold, R.~Cimrman, I.~Henriksen, E.~A. Quintero, C.~R. Harris, A.~M. Archibald, A.~H. Ribeiro, F.~Pedregosa, P.~van Mulbregt, and S.~1.0~Contributors, ``Scipy 1.0: fundamental algorithms for scientific computing in python,'' \emph{Nature Methods}, vol.~17, no.~3, pp. 261--272, Mar 2020. [Online]. Available: \url{https://doi.org/10.1038/s41592-019-0686-2}
\BIBentrySTDinterwordspacing

\bibitem{lehoucq1998arpack}
R.~B. Lehoucq, D.~C. Sorensen, and C.~Yang, \emph{ARPACK users' guide: solution of large-scale eigenvalue problems with implicitly restarted Arnoldi methods}.\hskip 1em plus 0.5em minus 0.4em\relax SIAM, 1998.

\bibitem{paszke2019pytorch}
A.~Paszke, S.~Gross, F.~Massa, A.~Lerer, J.~Bradbury, G.~Chanan, T.~Killeen, Z.~Lin, N.~Gimelshein, L.~Antiga \emph{et~al.}, ``Pytorch: An imperative style, high-performance deep learning library,'' \emph{Advances in neural information processing systems}, vol.~32, 2019.

\bibitem{lian2011shape}
Z.~Lian, A.~Godil, B.~Bustos, M.~Daoudi, J.~Hermans, S.~Kawamura, Y.~Kurita, G.~Lavoua, P.~D. Suetens \emph{et~al.}, ``Shape retrieval on non-rigid 3d watertight meshes,'' in \emph{Eurographics workshop on 3d object retrieval (3DOR)}.\hskip 1em plus 0.5em minus 0.4em\relax Citeseer, 2011.

\bibitem{pickup2014shrec}
D.~Pickup, X.~Sun, P.~L. Rosin, R.~Martin, Z.~Cheng, Z.~Lian, M.~Aono, A.~B. Hamza, A.~Bronstein, M.~Bronstein \emph{et~al.}, ``Shrec’14 track: Shape retrieval of non-rigid 3d human models,'' in \emph{Proceedings of the 7th Eurographics workshop on 3D Object Retrieval}, vol.~1, no.~2.\hskip 1em plus 0.5em minus 0.4em\relax Eurographics Association, 2014, p.~6.

\bibitem{10.5555/2852282.2852307}
Z.~Lian, J.~Zhang, S.~Choi, H.~ElNaghy, J.~El-Sana, T.~Furuya, A.~Giachetti, R.~A. Guler, L.~Lai, C.~Li, H.~Li, F.~A. Limberger, R.~Martin, R.~U. Nakanishi, A.~P. Neto, L.~G. Nonato, R.~Ohbuchi, K.~Pevzner, D.~Pickup, P.~Rosin, A.~Sharf, L.~Sun, X.~Sun, S.~Tari, G.~Unal, and R.~C. Wilson, ``Non-rigid 3d shape retrieval,'' in \emph{Proceedings of the 2015 Eurographics Workshop on 3D Object Retrieval}, ser. 3DOR '15.\hskip 1em plus 0.5em minus 0.4em\relax Goslar, DEU: Eurographics Association, 2015, p. 107–120.

\bibitem{yi2016scalable}
L.~Yi, V.~G. Kim, D.~Ceylan, I.-C. Shen, M.~Yan, H.~Su, C.~Lu, Q.~Huang, A.~Sheffer, and L.~Guibas, ``A scalable active framework for region annotation in 3d shape collections,'' \emph{ACM Transactions on Graphics (ToG)}, vol.~35, no.~6, pp. 1--12, 2016.

\bibitem{shilane2004princeton}
P.~Shilane, P.~Min, M.~Kazhdan, and T.~Funkhouser, ``The princeton shape benchmark,'' in \emph{Proceedings Shape Modeling Applications, 2004.}\hskip 1em plus 0.5em minus 0.4em\relax IEEE, 2004, pp. 167--178.

\bibitem{kanezaki2018rotationnet}
A.~Kanezaki, Y.~Matsushita, and Y.~Nishida, ``Rotationnet: Joint object categorization and pose estimation using multiviews from unsupervised viewpoints,'' in \emph{Proceedings of the IEEE conference on computer vision and pattern recognition}, 2018, pp. 5010--5019.

\bibitem{wei2020view}
X.~Wei, R.~Yu, and J.~Sun, ``View-gcn: View-based graph convolutional network for 3d shape analysis,'' in \emph{Proceedings of the IEEE/CVF conference on computer vision and pattern recognition}, 2020, pp. 1850--1859.

\bibitem{hamdi2021mvtn}
A.~Hamdi, S.~Giancola, and B.~Ghanem, ``Mvtn: Multi-view transformation network for 3d shape recognition,'' in \emph{Proceedings of the IEEE/CVF international conference on computer vision}, 2021, pp. 1--11.

\bibitem{limberger2017spectral}
F.~A. Limberger, ``Spectral signatures for non-rigid 3d shape retrieval,'' Ph.D. dissertation, University of York, 2017.

\bibitem{luciano2019global}
L.~Luciano and A.~B. Hamza, ``A global geometric framework for 3d shape retrieval using deep learning,'' \emph{Computers \& Graphics}, vol.~79, pp. 14--23, 2019.

\bibitem{bu2014learning}
S.~Bu, Z.~Liu, J.~Han, J.~Wu, and R.~Ji, ``Learning high-level feature by deep belief networks for 3-d model retrieval and recognition,'' \emph{IEEE Transactions on Multimedia}, vol.~16, no.~8, pp. 2154--2167, 2014.

\bibitem{lahav2020meshwalker}
A.~Lahav and A.~Tal, ``Meshwalker: Deep mesh understanding by random walks,'' \emph{ACM Transactions on Graphics (TOG)}, vol.~39, no.~6, pp. 1--13, 2020.

\bibitem{milano2020primal}
F.~Milano, A.~Loquercio, A.~Rosinol, D.~Scaramuzza, and L.~Carlone, ``Primal-dual mesh convolutional neural networks,'' \emph{Advances in Neural Information Processing Systems}, vol.~33, pp. 952--963, 2020.

\bibitem{mitchel2021field}
T.~W. Mitchel, V.~G. Kim, and M.~Kazhdan, ``Field convolutions for surface cnns,'' in \emph{Proceedings of the IEEE/CVF International Conference on Computer Vision}, 2021, pp. 10\,001--10\,011.

\bibitem{kim2021agcn}
S.~Kim and D.~C. Alexander, ``Agcn: Adversarial graph convolutional network for 3d point cloud segmentation,'' in \emph{32nd British Machine Vision Conference, BMVC 2021}.\hskip 1em plus 0.5em minus 0.4em\relax The British Machine Vision Association (BMVA), 2021, pp. 1--13.

\bibitem{xu2018spidercnn}
Y.~Xu, T.~Fan, M.~Xu, L.~Zeng, and Y.~Qiao, ``Spidercnn: Deep learning on point sets with parameterized convolutional filters,'' in \emph{Proceedings of the European Conference on Computer Vision (ECCV)}, 2018, pp. 87--102.

\bibitem{huang2018robust}
J.~Huang, H.~Su, and L.~Guibas, ``Robust watertight manifold surface generation method for shapenet models,'' \emph{arXiv preprint arXiv:1802.01698}, 2018.

\bibitem{maron2017convolutional}
H.~Maron, M.~Galun, N.~Aigerman, M.~Trope, N.~Dym, E.~Yumer, V.~G. Kim, and Y.~Lipman, ``Convolutional neural networks on surfaces via seamless toric covers.'' \emph{ACM Trans. Graph.}, vol.~36, no.~4, pp. 71--1, 2017.

\bibitem{wang2012active}
Y.~Wang, S.~Asafi, O.~Van~Kaick, H.~Zhang, D.~Cohen-Or, and B.~Chen, ``Active co-analysis of a set of shapes,'' \emph{ACM Transactions on Graphics (TOG)}, vol.~31, no.~6, pp. 1--10, 2012.

\bibitem{bogo2014faust}
F.~Bogo, J.~Romero, M.~Loper, and M.~J. Black, ``Faust: Dataset and evaluation for 3d mesh registration,'' in \emph{Proceedings of the IEEE conference on computer vision and pattern recognition}, 2014, pp. 3794--3801.

\bibitem{thomas2019kpconv}
H.~Thomas, C.~R. Qi, J.-E. Deschaud, B.~Marcotegui, F.~Goulette, and L.~J. Guibas, ``Kpconv: Flexible and deformable convolution for point clouds,'' in \emph{Proceedings of the IEEE/CVF international conference on computer vision}, 2019, pp. 6411--6420.

\bibitem{Wiersma2020}
K.~H. Ruben~Wiersma, Elmar~Eisemann, ``Cnns on surfaces using rotation-equivariant features,'' \emph{Transactions on Graphics}, vol.~39, no.~4, Jul. 2020.

\bibitem{cao2023unsupervised}
D.~Cao, P.~Roetzer, and F.~Bernard, ``Unsupervised learning of robust spectral shape matching,'' \emph{arXiv preprint arXiv:2304.14419}, 2023.

\bibitem{van2008visualizing}
L.~Van~der Maaten and G.~Hinton, ``Visualizing data using t-sne.'' \emph{Journal of machine learning research}, vol.~9, no.~11, 2008.

\end{thebibliography}

\end{document}